%% file: main.tex
\documentclass[letterpaper, 10pt, conference]{ieeeconf}      % Use this line for a4 paper

\IEEEoverridecommandlockouts                              % This command is only needed if 
\usepackage{amsmath,amsfonts}
\usepackage{amssymb}
\usepackage{amsthm}
\usepackage{algorithmic}
\usepackage{algorithm}
\usepackage{array}
\usepackage{amsmath}

\usepackage[caption=false,font=normalsize,labelfont=sf,textfont=sf]{subfig}
\usepackage{textcomp}
\usepackage{stfloats}
\usepackage{url}
\usepackage{verbatim}
\usepackage{graphicx}
\usepackage{fontawesome5}
\usepackage{cite}
\usepackage{soul}
\input{utils/mypackages}

\input{utils/mycommands}
\usepackage{yfonts}
\newtheorem{theorem}{Theorem}

\newtheorem{proposition}{Proposition}

\newtheorem{assumption}{Assumption}

\newtheorem{remark}{Remark}
\definecolor{darkteal}{HTML}{00565B}
\definecolor{citeblue}{HTML}{1A3FA0}
\definecolor{purple}{HTML}{3400A3}

\makeatletter
\newcommand{\customlabel}[2]{%
  \protected@edef\@currentlabel{#2}%
  \label{#1}%
}
\makeatother
\definecolor{mygreen}{rgb}{0.1,0.86,0.16}

\usepackage[pdftex, plainpages=false,hypertexnames=true,pdfnewwindow=true,colorlinks=true,citecolor=citeblue,linkcolor=citeblue,urlcolor=citeblue,filecolor=citeblue]{hyperref}

\begin{document}

\title{\LARGE \bf Learning Robot Metareasoning: Adaptive Computation Budgets for Real-Time Control
}

\author{
  Lorenzo Magnino$^1$, Antonio Marino$^1$, Michael Amir$^1$, Daigo Shishika$^2$, Amanda Prorok$^1$
  \thanks{This work is supported by European Research Council (ERC) Project 949940 (gAIa) and ARL DCIST CRA W911NF-17-2-0181. We gratefully acknowledge their support.} \thanks{$^1$L. Magnino, A. Marino, M. Amir, and A.Prorok are with University of Cambridge --- Cambridge, United Kingdom. Emails: \{lm2183, am3507,ma2151,asp45\}@cam.ac.uk. $^2$D. Shishika is with are with College of Engineering and Computing at George Mason University, Email: dshishik@gmu.edu}
}


\maketitle
\begin{abstract}
Planning while moving faces an historical trade-off between deliberation and reactivity. 
A planner running concurrently with execution needs time to compute, so the robot must commit to an open-loop action chunk for that duration: a longer commitment buys a better plan at the cost of feedback frequency. Existing controllers fix this action commitment, and consequently the computation budget, before deployment. However, the optimal computation time is highly state-dependent (due to factors such as unmodeled dynamics, safety constraints, and marginal optimisation gains), i.e. dynamic, therefore offline static tuning risks leaving systems blind during critical maneuvers or unable to find global effective solutions.

We cast the identification of the optimal computation budget as a semi-Markov decision process (SMDP) in which a single decision defines both the robot action commitments and the time allocated to the solver. To solve this, we train a reinforcement-learning policy conditioned on the current state, the executing plan, and inexpensive planner statistics. Across ground-robot navigation, agile quadrotor flight, and contact-rich manipulation, the learned policy changes its budget with the demands of the scenario, discovering efficient reasoning patterns that outperform fixed-budget baselines.
% We release additional details at 
\href{https://lorenzomagnino.github.io/learning-robot-metareasoning-review/}{\faGlobe\ website}
% \url{https://anonymous.4open.science/w/learning-robot-metareasoning-review-68DC/}.

\end{abstract}  

%\begin{IEEEkeywords}
%Robot Metareasoning, Robot Learning, Action Chunking
%\end{IEEEkeywords}

\input{chapters/introduction}

\input{chapters/related_work}
\input{chapters/preliminaries}
\input{chapters/surprise_as_closed_loop_feedback}
\input{chapters/experiments}
\input{chapters/conclusion}

\bibliography{bibliography/world_model,bibliography/rl_theory,bibliography/chunking,bibliography/libraries}
\bibliographystyle{IEEEtran}

% \input{chapters/appendix}

 

% \end{IEEEbiography}



\newpage
\end{document}

%% file: utils/mypackages.tex
\usepackage{multicol}
\usepackage{pgfplots}
\usepackage{lipsum}
\usepackage{amsfonts,amsmath,amssymb,mathtools,bbm,dsfont,stmaryrd,aligned-overset,bm}
\usepackage{graphicx}
\usepackage{wrapfig}
\usepackage{epstopdf}
\usepackage{amsopn}
\usepackage{cancel}
\usepackage{physics}
\usepackage[framemethod=TikZ]{mdframed}
\usepackage{subcaption}
\usepackage{comment}
\usepackage{booktabs}
\usepackage{colortbl}
\definecolor{gapteal}{HTML}{00665E}
\definecolor{gapblue}{HTML}{247BB5}
\definecolor{gappurple}{HTML}{512579}
\definecolor{gaprose}{HTML}{C05078}
\usepackage{physics}
\usepackage{microtype}
\let\labelindent\relax
\usepackage{enumitem}\setlist[enumerate,1]{label={(\roman*)}}
\ifpdf
  \DeclareGraphicsExtensions{.eps,.pdf,.png,.jpg}
\else
  \DeclareGraphicsExtensions{.eps}
\fi

\newbool{arxiv}
\boolfalse{arxiv}

\usepackage{multirow}

\usepackage{tikz,pgfplots}
\usetikzlibrary{shapes.geometric, arrows.meta, positioning, fit, calc, backgrounds, shadows, patterns,hobby,calc,intersections, decorations.pathmorphing, decorations.pathreplacing, decorations.markings, fadings, fit, backgrounds}

\tikzset{
    antibody/.style={
        circle,
        draw=blue!10,      % Very faint border
        fill=blue!5,       % Very faint background
        minimum size=1.6cm,
        path picture={
            \draw[line width=3.5pt, #1, line cap=round, line join=round] 
            (0,-0.35) -- (0,0) -- (-0.35,0.4) 
            (0,0) -- (0.35,0.4);
        }
    },
    virus/.style={
        circle, 
        fill=#1!20,        % Soft pastel interior
        draw=#1!90!black,  % Darker outline
        line width=1.2pt,
        minimum size=1.1cm,
        path picture={
            \fill[#1!30] (0,0) circle (0.35cm);
            \foreach \a in {0,45,...,315} {
                \draw[line width=1.5pt, #1!90!black, line cap=round] 
                (\a:0.35) -- (\a:0.55); % The stalk
                \fill[#1!90!black] (\a:0.6) circle (1.5pt); % The receptor head
            }
        }
    },
    label text/.style={
        font=\sffamily\small, 
        align=center,
        text=black!80
    },
    math text/.style={
        font=\scriptsize\bfseries,
        text=black!50
    }
}

\tikzset{
    vessel/.style={
        draw=gray!80,
        line width=2pt,
        fill=gray!10,
        rounded corners=5pt
    },
    coil/.style={
        rectangle,
        draw=orange!80!black,
        fill=orange!40,
        minimum width=0.4cm,
        minimum height=0.25cm,
        rounded corners=2pt
    },
    plasma/.style={
        draw=#1!80!black,
        fill=#1!60,
        opacity=0.8,
        line width=1.5pt,
        line join=round,
        line cap=round
    },
    label text/.style={
        font=\sffamily\small, 
        align=center,
        text=black!80
    },
    math text/.style={
        font=\scriptsize\bfseries,
        text=black!60
    }
}

\pgfplotsset{compat=1.16}
\usepackage{pgfplotstable}
\usepgfplotslibrary{fillbetween}

\usepackage{pdfpages}

\usepackage{centernot}

\usepackage[acronym]{glossaries}
\newmdenv[
  topline=false,
  bottomline=false,
  rightline=false,
  linewidth=3pt,        % Thickness of the line
  linecolor=gray,% Color of the line
  innerleftmargin=10pt, % Distance between line and text
  backgroundcolor=gray!10,  % The background color (requires xcolor)
  innertopmargin=15pt,       % Padding at the top
  innerbottommargin=10pt,    % Padding at the bottom
  innerrightmargin=5pt,     % Padding on the right
  skipabove=10pt,
  skipbelow=10pt
]{leftbar}

\usepackage{tcolorbox}
\newtcolorbox{myquote}{
  blanker,
  borderline west={3pt}{0pt}{gray},
  left=10pt,
  top=4pt,
  bottom=4pt
}

%% file: utils/mycommands.tex
\newbool{proofInMain}
\newbool{proofInAppendix}
\newbool{duplicateStatementInAppendix}
\newbool{showLinks}
\newbool{showBoxes}

\boolfalse{proofInMain}
\booltrue{proofInAppendix}
\boolfalse{duplicateStatementInAppendix}
\boolfalse{showLinks}
\boolfalse{showBoxes}

\newcommand{\WikipediaLink}[2]{\ifbool{showLinks}{\href{#1}{\color{blue}{#2}}}{#2}}%
\newcommand{\StackExchangeLink}[2]{\ifbool{showLinks}{\href{#1}{\color{purple}{#2}}}{#2}}%
\newcommand{\emphBox}[2]{%
\ifbool{showBoxes}{%
\begin{tcolorbox}[boxrule=0pt, frame empty, colback=lightgray, breakable]
\begin{center}
\textbf{#1}
\end{center}
#2
\end{tcolorbox}
}{%
#2}}

\newcommand{\localSubseteq}[1]               {\subseteq} %_{#1}

\input{utils/tikzart}

%% file: chapters/introduction.tex
\section{Introduction}

\label{sec:pvoc}

Deciding \emph{how much} to think is as hard as the thinking itself (e.g., deciding
what to do)~\cite{otto2019opportunity}. Biological agents dynamically arbitrate their cognitive resources, continuously balancing rapid, reactive responses against slower, deliberate reasoning based on environmental uncertainty and temporal constraints~\cite{otto2019opportunity, daw2005uncertainty}. This adaptive allocation represents a fundamental speed-optimality trade-off where extended deliberation is penalized by elapsed time~\cite{keramati2011speed, shenhav2017toil}.
%\paragraph*{Economical and physical value of computation} 
The \emph{value of computation} (VOC) formally characterizes this trade-off between the marginal gain in decision optimality and the cost of deliberation~\cite{russell1991metareasoning,hay2012selecting}. Traditional metareasoning frameworks assume a frozen world state during planning, requiring an explicitly defined cost function to penalize extended computation, a paradigm we refer to as the \emph{economic} value of computation. Embodied autonomous agents, conversely, operate in continuous time where this assumption fails. Because the robot and its environment evolve concurrently with the planner's execution, heuristic timing penalties are unnecessary. Instead, the physical dynamics inherently penalize computational delay through unmitigated state drift, thereby hindering high-frequency reactivity and reducing safety margins. We introduce this dynamic trade-off as the \emph{physical} value of computation, formalized in Sec.~\ref{subsec:smdp}.

%Autonomous robotic systems operating in dynamic environments face an analogous arbitration. A physical planner must balance the theoretical optimality of a deeply deliberated, long-horizon trajectory against the closed-loop robustness provided by high-frequency reactive control. Yet, unlike biological systems, modern robotic controllers typically operate with rigid computational budgets, lacking the architectural flexibility to adapt their reasoning horizon to the immediate physical constraints and dynamic evolution of the environment.

\begin{figure}[t!]
    \centering
    \tikzfading[name=rollout fade,top color=transparent!70,bottom color=transparent!0]
    \tikzfading[name=rollout horizontal fade,left color=transparent!0,right color=transparent!65]
    \makebox[\linewidth][l]{%
    \begin{tikzpicture}
      \node[anchor=south west,inner sep=0,outer sep=0] (real) at (-7pt,0)
        {\includegraphics[width=0.31\linewidth,height=0.27\linewidth,trim=120bp 19.56129bp 175bp 55.6bp,clip]{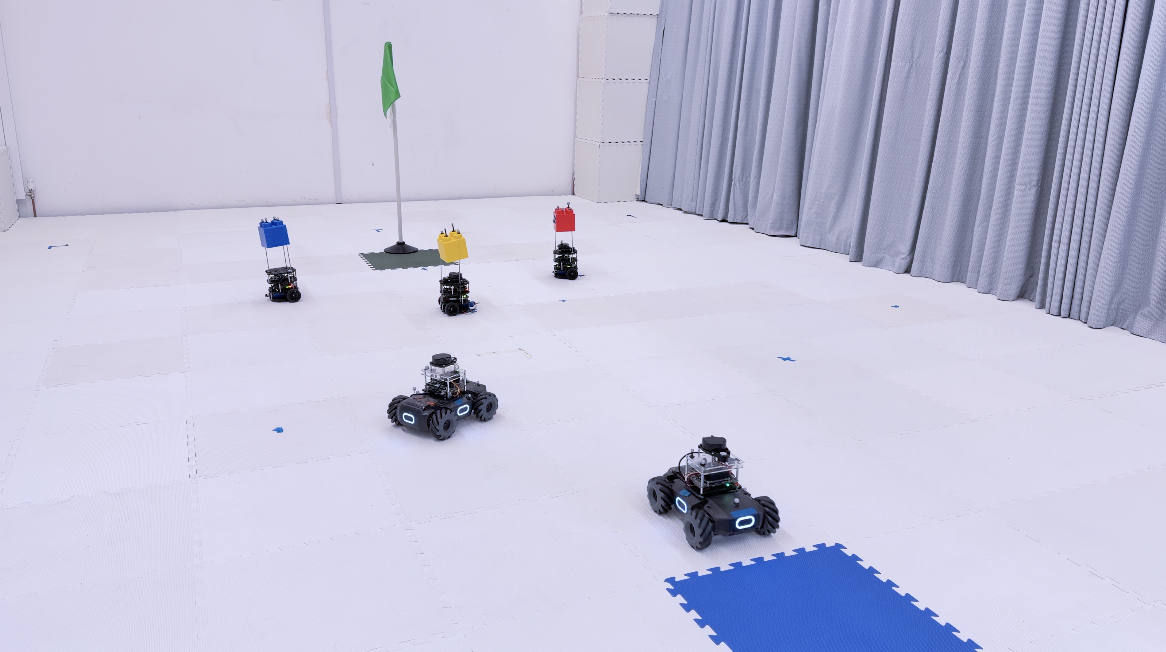}};
      \node[anchor=south west,inner sep=0,outer sep=0] (highway) at (real.south east)
        {\includegraphics[width=0.31\linewidth,height=0.27\linewidth,trim=0 0 200bp 0,clip]{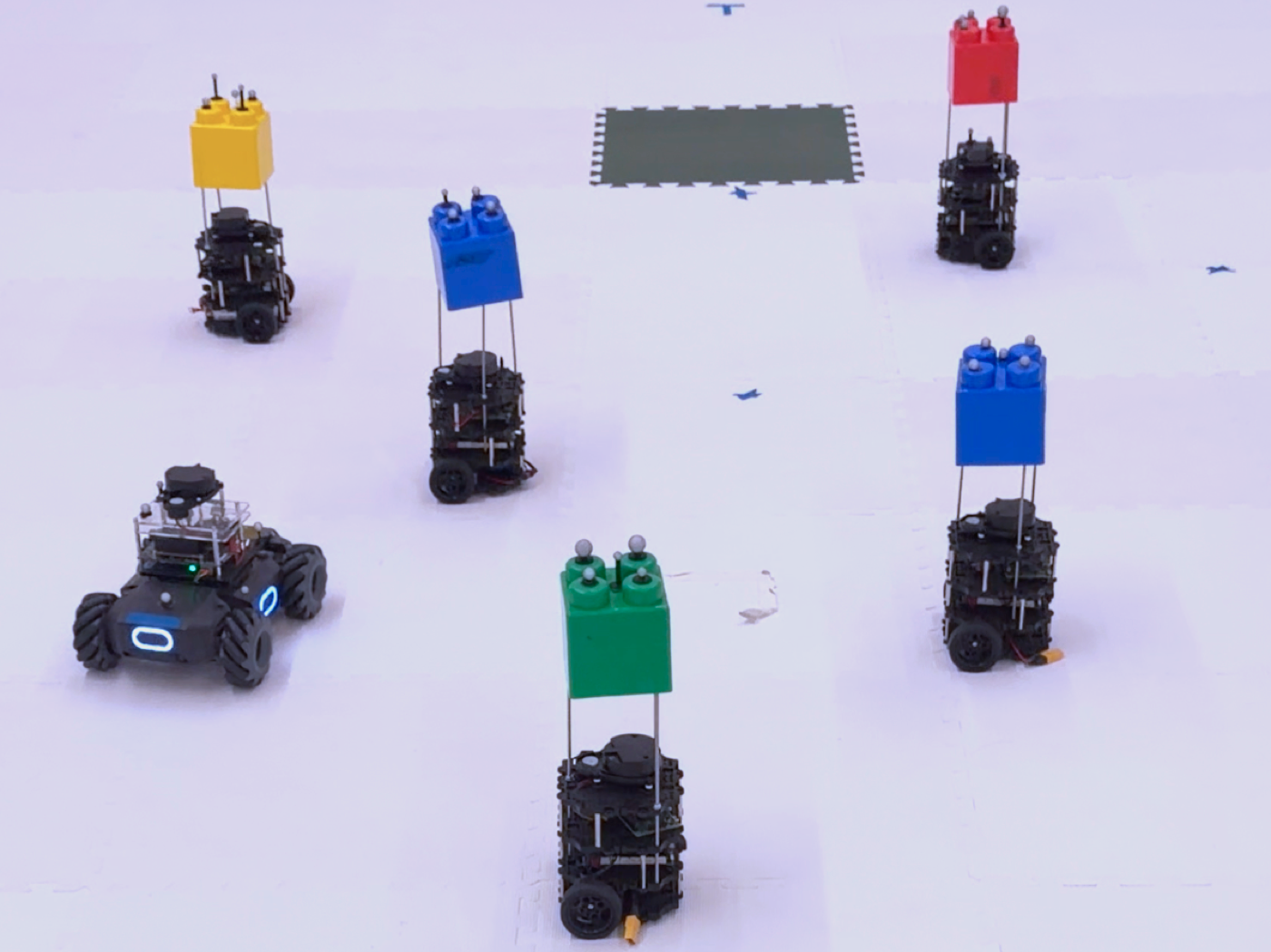}};
      \node[anchor=south west,inner sep=0,outer sep=0] (barrier) at (highway.south east)
        {\includegraphics[width=0.31\linewidth,height=0.27\linewidth,trim=78.75bp 0 113.75bp 0,clip]{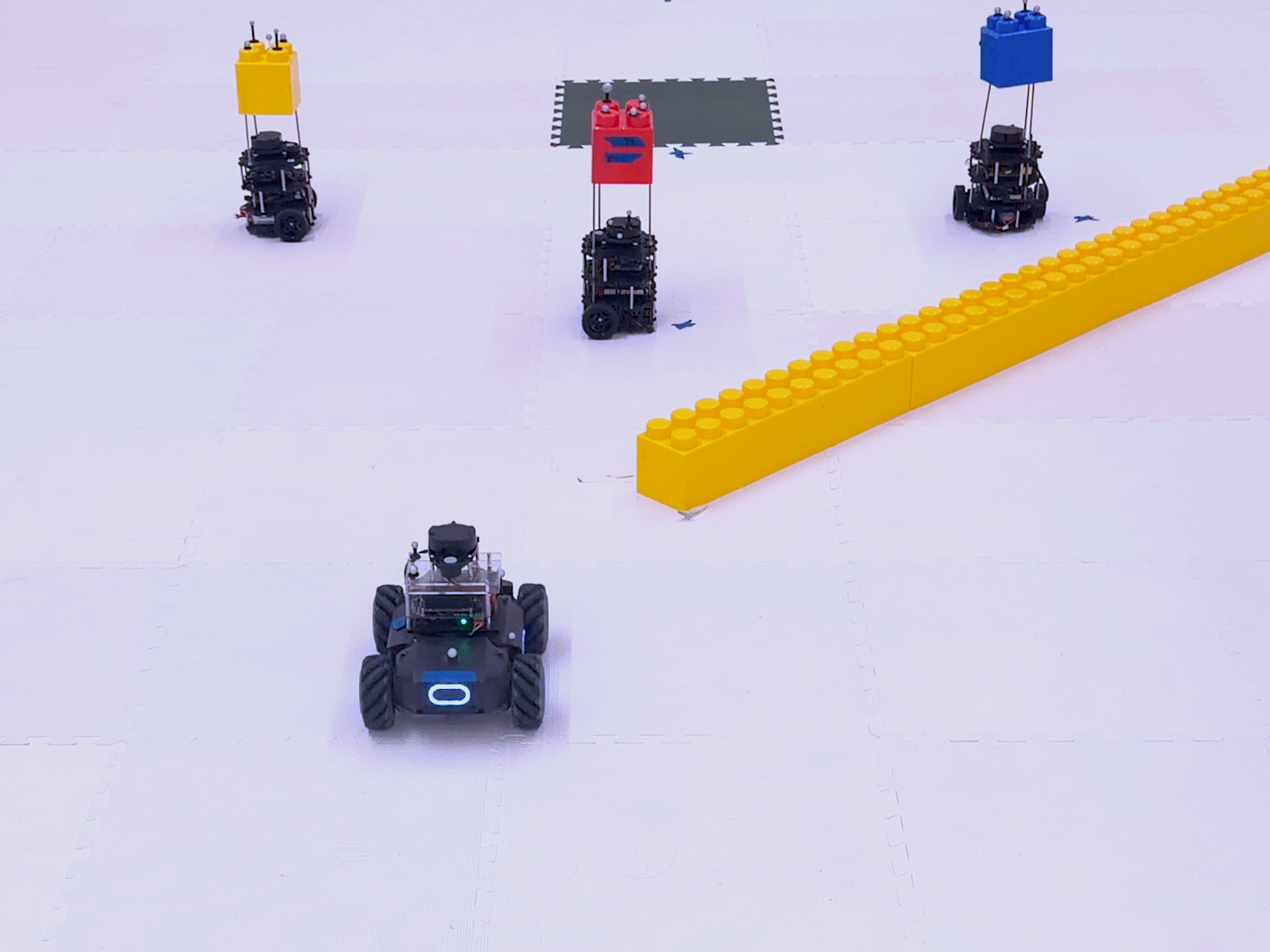}};
      \definecolor{heroNavy}{HTML}{142D4E}
      \definecolor{heroRed}{HTML}{C74343}
      \definecolor{budgetSalmon}{HTML}{FA8072}
      % Bright magnified ego views over a lightly dimmed scene.
      \foreach \panel/\asset/\trim/\egoX/\egoY/\zoom/\spotRadius in {
        real/hero_real_navigation/{120bp 19.56129bp 175bp 55.6bp}/.823/.239/1.3/.06,
        highway/hero_highway_matched/{0 0 200bp 0}/.183/.395/1.0/.055,
        barrier/hero_barrier_matched/{78.75bp 0 113.75bp 0}/.349/.333/1.3/.06} {
        \begin{scope}[shift={(\panel.south west)},x=0.31\linewidth,y=0.27\linewidth]
          \clip (0,0) rectangle (1,1);
          \fill[black!70,opacity=.15] (0,0) rectangle (1,1);
          \pgfmathsetlengthmacro{\zoomwidth}{.31*\zoom*\linewidth}
          \pgfmathsetlengthmacro{\zoomheight}{.27*\zoom*\linewidth}
          \edef\zoomimage{\noexpand\includegraphics[width=\zoomwidth,height=\zoomheight,trim=\trim,clip]{figures/assets/\asset.png}}
          \begin{scope}
            \clip (\egoX,\egoY) circle[radius=\spotRadius\linewidth];
            \node[anchor=south west,inner sep=0,outer sep=0]
              at ({(1-\zoom)*\egoX},{(1-\zoom)*\egoY}) {\zoomimage};
          \end{scope}
          \draw[white,line width=.6pt] (\egoX,\egoY) circle[radius=\spotRadius\linewidth];
        \end{scope}
      }
      \tikzset{motion/.style={-{Stealth[length=2.6pt,width=2.4pt]},line width=.8pt,draw=black!65,opacity=.55}}
      % Motion overlays use coordinates relative to each navigation panel.
      \begin{scope}[shift={([yshift=0.022\linewidth]real.south west)},x={(0.31\linewidth,0)},y={(0,0.248\linewidth)}]
        \clip (0,-0.08870968) rectangle (1,1);
        \foreach \x/\y in {.330/.596,.115/.645,.601/.699}
          \fill[black!70,opacity=.65] (\x,\y) circle[radius=.6pt];
        \draw[motion] (.330,.596) -- (.255,.596);
        \draw[motion] (.115,.645) -- (.180,.645);
        \draw[motion] (.601,.699) -- (.671,.699);
      \end{scope}
      \begin{scope}[shift={(highway.south west)},x={(0.36166667\linewidth,0)},y={(0,0.27\linewidth)}]
        \clip (0,0) rectangle (.85714286,1);
        \foreach \x/\y in {0.264/.700,0.447/.540,0.711/.346,0.551/.148}
          \fill[black!70,opacity=.65] (\x,\y) circle[radius=.6pt];
        \fill[black!70,opacity=.5] (0.711,.751) circle[radius=.6pt];
        \draw[motion] (0.264,.700) -- (0.320,.700);
        \draw[motion,line width=1pt,opacity=.42] (0.711,.751) -- (0.632,.751);
        \draw[motion] (0.447,.540) -- (0.519,.540);
        \draw[motion] (0.711,.346) -- (0.632,.346);
        \draw[motion] (0.551,.148) -- (0.641,.148);
      \end{scope}
      \begin{scope}[shift={([xshift=-0.02021739\linewidth]barrier.south west)},x={(0.33695652\linewidth,0)},y={(0,0.27\linewidth)}]
        \clip (.06,0) rectangle (.98,1);
        \foreach \x/\y in {0.281/.794,0.766/.794,0.581/.700}
          \fill[black!70,opacity=.65] (\x,\y) circle[radius=.6pt];
        \draw[motion] (0.281,.794) -- (0.338,.794);
        \draw[motion] (0.766,.794) -- (0.693,.794);
        \draw[motion] (0.581,.700) -- (0.640,.700);
      \end{scope}
      \definecolor{rolloutBlue}{HTML}{529ADD}
      \definecolor{rolloutIndigo}{HTML}{7388D8}
      \definecolor{rolloutViolet}{HTML}{9272D1}
      \definecolor{rolloutPurple}{HTML}{AC61C8}
      \tikzset{sampled rollout/.style={-{Straight Barb[length=1.3pt,width=1.4pt,round]},
        line width=0.8pt,line cap=round,line join=round,path fading=rollout fade,
        decoration={markings,mark=between positions 3pt and -\pgfkeysvalueof{/tikz/rollout marker end} step 3pt with
          {\pgfmathsetmacro{\fadeProgress}{max(0,(\pgfkeysvalueof{/pgf/decoration/mark info/distance from start}/\pgfdecoratedpathlength-\pgfkeysvalueof{/tikz/rollout marker fade start})/(1-\pgfkeysvalueof{/tikz/rollout marker fade start})))}
           \pgfmathsetmacro{\stateOpacity}{\pgfkeysvalueof{/tikz/rollout marker opacity}*(1-\pgfkeysvalueof{/tikz/rollout marker fade}*\fadeProgress)}
           \fill[fill=\pgfkeysvalueof{/tikz/rollout marker color},opacity=\stateOpacity] (0,0) circle[radius=\pgfkeysvalueof{/tikz/rollout marker radius}];}}},
        rollout marker color/.initial=rolloutBlue,
        rollout marker opacity/.initial=.7,
        rollout marker fade/.initial=.7,
        rollout marker fade start/.initial=0,
        rollout marker end/.initial=2pt,
        rollout marker radius/.initial=.45pt,
        rollout color/.style={color=#1,rollout marker color=#1,
          postaction={draw opacity=0,fill opacity=1,path fading=none,decorate}}}
      \tikzset{ego rollout/.style={sampled rollout,arrows=-,
        line width=1pt,rollout color=heroRed,rollout marker radius=.85pt,
        rollout marker opacity=1,rollout marker fade=.9,rollout marker fade start=.35,opacity=1},
        ego arrowhead/.style={draw=heroRed,draw opacity=0,fill opacity=1,line width=0pt,
          -{Stealth[length=2.6pt,width=2.4pt]}}}
      \begin{scope}[shift={(real.south west)},x=0.31\linewidth,y=0.27\linewidth]
        \fill[black!70,opacity=.65] (.440,.470) circle[radius=.6pt];
        \draw[motion] (.440,.470) -- (.686,.610);
      \end{scope}
      % Illustrative rollout fans traced from the reference, not recorded trajectories.
      \begin{scope}[shift={([yshift=0.022\linewidth]real.south west)},x={(0.31\linewidth,0)},y={(0,0.248\linewidth)}]
        \clip (0,-0.08870968) rectangle (1,1);
        \draw[ego rollout,rollout marker end=5pt] (0.709,0.251) .. controls (0.612,0.313) and (0.390,0.282) .. (0.307,0.319) .. controls (0.254,0.343) and (0.252,0.456) .. (0.247,0.557);
        \draw[ego arrowhead] (0.252,0.456) -- (0.246,0.575);
        \fill[heroRed] (0.709,0.251) circle[radius=.9pt];
      \end{scope}
      \begin{scope}[shift={(highway.south west)},x={(0.36166667\linewidth,0)},y={(0,0.27\linewidth)}]
        \clip (0,0) rectangle (.85714286,1);
        \draw[ego rollout] (0.221,0.477) .. controls (0.245,0.513) and (0.184,0.575) .. (0.231,0.604);
        \draw[ego arrowhead] (0.195,0.575) -- (0.245,0.614);

        \fill[heroRed] (0.221,0.477) circle[radius=.9pt];
      \end{scope}
      \begin{scope}[shift={([xshift=-0.02021739\linewidth]barrier.south west)},x={(0.33695652\linewidth,0)},y={(0,0.27\linewidth)}]
        \clip (.06,0) rectangle (.98,1);
        \begin{scope}[yshift=.012\linewidth]

        \draw[ego rollout] (0.371,0.477) .. controls (0.332,0.558) and (0.314,0.660) .. (0.353,0.733);
        \draw[ego arrowhead] (0.314,0.660) -- (0.361,0.749);

        \fill[heroRed] (0.371,0.477) circle[radius=.9pt];
        \end{scope}
      \end{scope}

      \tikzset{hero label/.style={anchor=north west,inner sep=0,outer sep=0,
        minimum height=14pt,text width=0.31\linewidth,align=center,
        font=\scriptsize,text=black,fill=white}}
      \node[hero label] (real-label) at ([yshift=-3pt]real.south west)
        {Deliberative\quad {\fontsize{7}{7}\selectfont\textcolor{gappurple}{\faBrain}}\raisebox{2pt}{\fontsize{6}{6}\selectfont\textcolor{gappurple}{\faClock}}};
      \node[hero label] (highway-label) at ([yshift=-3pt]highway.south west)
        {Reactive\quad {\fontsize{5}{5}\selectfont\textcolor{gappurple}{\faBrain}}\raisebox{2pt}{\fontsize{6}{6}\selectfont\textcolor{gappurple}{\faBolt}}};
      \node[hero label] (barrier-label) at ([yshift=-3pt]barrier.south west)
        {Intermediate\quad {\fontsize{6}{6}\selectfont\textcolor{gappurple}{\faBrain}}};

      \foreach \panel/\filled/\state/\blockShift in {real/6/s/9,highway/1/s'/6,barrier/4/s''/9} {
        \begin{scope}[shift={([xshift=\blockShift pt,yshift=26pt]\panel.north)},x=1pt,y=1pt]
          \node[anchor=east,inner sep=0,text=black,font=\scriptsize,text height=5pt,text depth=1pt]
            at (-21.125,-13) {$\state\!\mapsto$};
          \foreach \i in {1,...,7} {
            \ifnum\i>\filled
              \def\fuelcolor{white}
            \else
              \pgfmathtruncatemacro{\salmonpercent}{100*(7-\i)/6}
              \edef\fuelcolor{budgetSalmon!\salmonpercent!gappurple}
            \fi
            \draw[draw=black!50,line width=.25pt,fill=\fuelcolor]
              ({-19.125+6.75*(\i-1)},-16) rectangle ++(6,6);
          }
          \path (0,-20);
        \end{scope}
      }
      \path ([xshift=-15pt]real.south west);
      \node[overlay,anchor=west,inner sep=0,text=black,font=\fontsize{6}{6.5}\selectfont\bfseries]
        at ([xshift=-21pt,yshift=13pt]real.north west) {BUDGET:};
      \foreach \panel in {real,highway}
        \draw[overlay,draw=gappurple,opacity=.4,line width=0.4pt] (\panel.south east) -- (\panel.north east);
      \foreach \panel/\state in {real/s,highway/s',barrier/s''}
        \node[anchor=north west,inner sep=1pt,outer sep=0,font=\scriptsize\bfseries,
          minimum width=12pt,minimum height=10pt,text height=6pt,text depth=1pt,align=center,
          text=gappurple,fill=white,fill opacity=.85,text opacity=1]
          at ([xshift=2pt,yshift=-2pt]\panel.north west) {$\state$};
    \end{tikzpicture}%
    }
    \captionsetup{skip=3pt}
    \caption{\small{\textbf{Robot Metareasoning: \textit{different states require different computation budgets}.}
    % Across real-world and simulated experiments, we show that adapting the computation budget across states within an episode is a fundamental capability for embodied intelligence.
    Robots must adapt their computation budget based on the situation that they are facing: ($s$) deliberate longer to search for an optimal path, ($s'$) react rapidly to noisy moving obstacles, or ($s''$) balance planning and reactivity. Across real-world and simulated experiments, we show that adapting the computation budget across states within an episode is a fundamental capability for embodied intelligence.}}
    \label{fig:hero}
    \vspace{-0.3cm}
\end{figure}
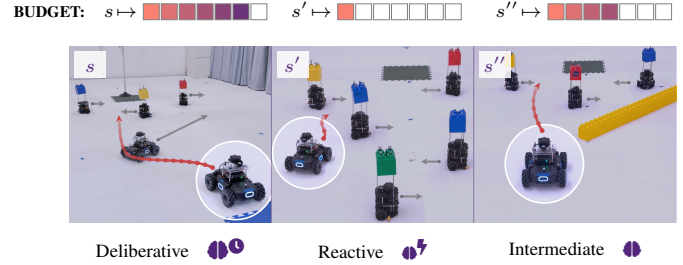

As standard for computational efficiency, robots typically plan in receding-horizon: the system optimizes a trajectory over a finite horizon, executes the initial segment, and subsequently replans from the evolved state. Model predictive control (MPC) \cite{williams2017mppi,bharadhwaj2020cem} commits to a single control step, resolving the problem at every control period, providing high-rate feedback around a
nominal trajectory. 

Alternatively, action-chunking architectures generate a sequence of actions
to execute in open-loop while the next plan is produced. This commitment can improve motion smoothness and sample efficiency \cite{zhao2023act,chi2023diffusion,black2024pi0}, yet fundamentally limits the system's ability to react to high-frequency disturbances or environmental changes. In both paradigms, the solver's computational budget and execution horizon are statically defined prior to deployment. Consequently, the robot \textit{lacks a principled mechanism to dynamically arbitrate when a state warrants extended deliberation for trajectory optimality versus when it necessitates rapid, reactive feedback (Fig.\ref{fig:hero})}.

Identifying an optimal static computation budget cannot be resolved via offline tuning. It is fundamentally a state-dependent quantity,  dictated by the unmodelled \emph{dynamics}, the proximity to state-space \emph{constraints}, and the marginal \emph{optimality} gained through additional solver iterations. Moreover, the budget choice cannot be assessed from the current state alone: its consequences propagate to future states throughout the episode, where they may influence later decisions and outcomes.

\textit{We address these challenges} by formulating adaptive computation budgets for real-time control
as a semi-MDP over decision epochs. At each epoch, one decision sets both the
open-loop execution horizon and the parallel computational resources allocated to solver. We then learn this decision from experience with an RL policy
(Fig.~\ref{fig:overview}).

Our \textbf{contributions} are the following:

\begin{figure*}[t]
    \centering
    \begin{minipage}{\textwidth}
    % \begin{minipage}[c]{0.30\textwidth}
    %     \centering
    %     \resizebox{\linewidth}{!}{\input{tikz/baselines}}
    % \end{minipage}\hspace{0.005\textwidth}%
    \begin{minipage}[c]{\textwidth}
        \centering
        \resizebox{0.90\linewidth}{!}{\input{tikz/smdp_pipeline}}
    \end{minipage}
    \end{minipage}
    \caption{\small 
    % Fixed versus adaptive commitment. \textbf{Left:} real-time
    % MPC commits to one step ($h=1$), whereas action chunking executes a fixed
    % block of $h$ actions. \textbf{Right:} 
    {\textbf{Method Overview}. At each decision epoch, the state, the previous plan and its statistics inform the budget policy $\Phi_\theta$, which selects the next execution horizon and thereby allocates the solver's budget. This lets the robot decide how much to deliberate under environmental uncertainty and temporal constraints. The new plan takes over, the state and statistics are refreshed, and the loop repeats.}}
    \label{fig:overview}
    \label{fig:baselines}
    \label{fig:smdp}
\end{figure*}
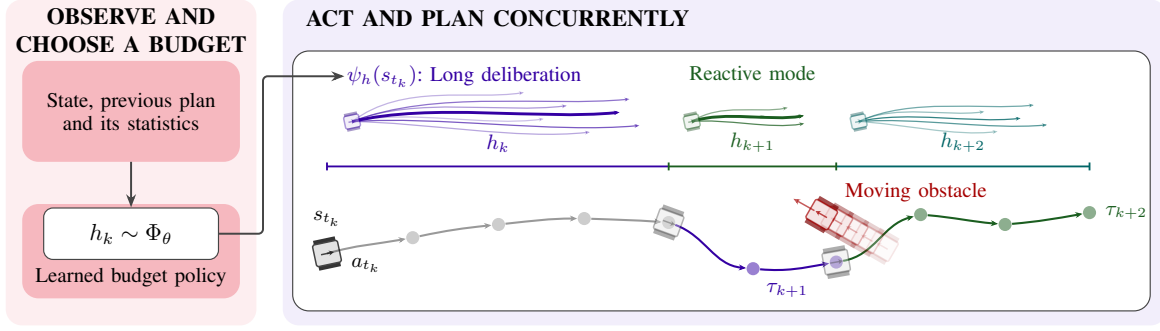

\begin{mdframed}[
    linecolor=blue!20,      % Sets the border color
    linewidth=0pt,        % Sets the border thickness
    roundcorner=5pt,        % Optional: rounds the corners slightly
    % hidealllines=true,
    backgroundcolor=blue!5,
    innertopmargin=8pt,
    innerbottommargin=8pt
]
\textit{(i)} \emph{Semi-MDP formulation.} We cast variable-computation adaptive chunking as a semi-MDP, in which the robot chooses how long to commit to a plan before replanning, and the solver budget follows from that choice.\\
\textit{(ii)} \emph{Adaptive policy through experience.} We propose an RL policy $\Phi_\theta$ that adapts the computation budget online, letting the controller trade reactivity against deeper solver optimisation on the fly. \\
\textit{(iii)} \emph{Insights from experiments.} Across different real-world and simulated environments, {we show that planner statistics are strong signals to the policy and that adaptive computation budgets match or outperform fixed-budget strategies across several robotic systems}.

\end{mdframed}

%% file: tikz/smdp_pipeline.tex
% Budget selection followed by concurrent execution and planning.
% Requires tikz; libraries arrows.meta, calc.
\begingroup
\definecolor{loopRose}{HTML}{F6B8BF}
\definecolor{loopLavender}{HTML}{C9B8F0}
\definecolor{loopInk}{HTML}{454545}
\definecolor{loopGreen}{HTML}{1B5E20}
\definecolor{loopTeal}{HTML}{006566}
\begin{tikzpicture}[
    x=1cm,y=1cm, font=\normalsize, text=black,
    flow/.style={-{Stealth[length=2mm,width=1.5mm]},
                 draw=loopInk,line width=.85pt,rounded corners=2pt},
    region/.style={rounded corners=7pt},
    card/.style={rounded corners=4pt,draw=loopInk,line width=.5pt,
                 fill=white,align=center,inner sep=5pt},
    heading/.style={font=\normalsize\bfseries},
    note/.style={font=\small,align=center},
    pics/robot/.style args={#1}{code={
        \fill[#1!75!black] (-.40,-.55) rectangle (.40,-.35);
        \fill[#1!75!black] (-.40,.35) rectangle (.40,.55);
        \draw[draw=#1,fill=#1!15,line width=.65pt,rounded corners=1pt]
            (-.46,-.40) rectangle (.46,.40);
        \draw[draw=#1,line width=.65pt,-{Stealth[length=2pt,width=2pt]}]
            (-.22,0) -- (.25,0);
    }},
]
% A vertical observation-and-budget column precedes the planning window.
\fill[region,loopRose!24] (-3.75,-.96) rectangle (.05,3.98);
\node[heading,align=center] at (-1.85,3.51) {OBSERVE AND\\CHOOSE A BUDGET};
\fill[region,loopRose] (-3.50,1.52) rectangle (-.20,3.05);
\fill[region,loopRose] (-3.50,-.55) rectangle (-.20,.88);
\node[note] at (-1.85,2.285) {State, previous plan\\and its statistics};
\node[card,minimum width=2.65cm,minimum height=.78cm] (policy) at (-1.85,.43)
    {$h_k\sim\Phi_\theta$};
\node[note] at (-1.85,-.23) {Learned budget policy};
\draw[flow] (-1.85,1.52) -- node[note,right=2pt] {} (policy.north);

% Adapted from overview.tex: execution and computation share one window.
\fill[region,loopLavender!24] (.45,-.96) rectangle (13.85,3.98);
\node[heading,anchor=west] at (.68,3.69) {ACT AND PLAN CONCURRENTLY};
\draw[card,rounded corners=7pt] (.60,-.74) rectangle (13.60,3.20);

% Upper segments mark consecutive commitment durations.
\draw[purple,line width=.8pt] (1.12,1.45) -- (6.30,1.45);
\draw[loopGreen,line width=.8pt] (6.30,1.45) -- (8.84,1.45);
\draw[loopTeal,line width=.8pt] (8.84,1.45) -- (12.68,1.45);
\draw[loopTeal,line width=.8pt] (12.68,1.39) -- (12.68,1.51);
\node[note,text=loopTeal,anchor=south] at (10.76,1.55) {$h_{k+2}$};
\draw[purple,line width=.8pt] (1.12,1.39) -- (1.12,1.51);
\foreach \x in {6.30,8.84}
    \draw[loopGreen,line width=.8pt] (\x,1.39) -- (\x,1.51);
\node[note,text=purple,anchor=south] at (3.71,1.55) {$h_k$};
\node[note,text=loopGreen,anchor=south] at (7.57,1.55) {$h_{k+1}$};
\node[note,text=purple,anchor=north,inner sep=1pt] (planner) at (3.20,3.02)
    {$\psi_h(s_{t_k})$: Long deliberation};
\node[note,text=loopGreen,anchor=north,inner sep=1pt] at (7.57,3.02)
    {Reactive mode};
\node[note,text=loopTeal,anchor=north,inner sep=1pt] at (10.76,3.02)
    {};
\draw[flow] (policy.east) -- (.15,.43) |- (planner.west);

% Schematic rollouts: the shorter next budget permits fewer, shorter searches.
\foreach \dy/\len/\alpha in {-.30/3.60/.35,-.20/4.35/.65,-.10/2.90/.25,0/4.05/.80,.10/3.25/.40,.20/4.20/.55,.30/2.60/.30} {
    \draw[purple,opacity=\alpha,line width=.55pt,-{Stealth[length=1.8pt,width=1.4pt]}]
        (1.50,2.13) .. controls ({1.50+.30*\len},{2.42+\dy})
        and ({1.50+.70*\len},{2.15+\dy}) .. ({1.50+\len},{2.27+\dy});
}
% Highlighted rollouts become the chunks handed to the next planning epoch.
\draw[purple,line width=1.25pt,-{Stealth[length=2.2pt,width=1.7pt]}]
    (1.50,2.13) .. controls (2.715,2.42) and (4.335,2.15) .. (5.55,2.27);
\pic[scale=.25,rotate=12,opacity=.40] at (1.50,2.13) {robot=purple};
\foreach \dy/\len/\alpha in {-.12/1.70/.75,0/1.10/.30,.12/1.45/.50} {
    \draw[loopGreen,opacity=\alpha,line width=.55pt,-{Stealth[length=1.8pt,width=1.4pt]}]
        (6.65,2.13) .. controls ({6.65+.30*\len},{2.30+\dy})
        and ({6.65+.70*\len},{2.13+\dy}) .. ({6.65+\len},{2.22+\dy});
}
\draw[loopGreen,line width=1.25pt,-{Stealth[length=2.2pt,width=1.7pt]}]
    (6.65,2.13) .. controls (7.16,2.30) and (7.84,2.13) .. (8.35,2.22);
\pic[scale=.25,rotate=12,opacity=.40] at (6.65,2.13) {robot=loopGreen};
\foreach \dy/\len/\alpha in {-.20/2.10/.35,-.10/2.85/.65,0/2.45/.80,.10/1.90/.40,.20/2.65/.55} {
    \draw[loopTeal,opacity=\alpha,line width=.55pt,-{Stealth[length=1.8pt,width=1.4pt]}]
        (9.18,2.13) .. controls ({9.18+.30*\len},{2.30+\dy})
        and ({9.18+.70*\len},{2.10+\dy}) .. ({9.18+\len},{2.18+\dy});
}
\pic[scale=.25,rotate=12,opacity=.40] at (9.18,2.13) {robot=loopTeal};
\begin{scope}[yshift=-.95cm]
\begin{scope}[
    dot/.style={circle,fill=black,inner sep=2.0pt,fill opacity=.50},
    chunkarrow/.style={-{Stealth[length=3.2pt,width=2.6pt]},
                 draw=black,line width=.9pt},
]
\coordinate (s0) at (1.12,1.08);
\pic[scale=.42,rotate=12] at (s0) {robot=loopInk};
\begin{scope}[opacity=.40,transparency group]
\node[dot] (s1) at (2.42,1.32) {};
\node[dot] (s2) at (3.72,1.52) {};
\node[dot] (s3) at (5.02,1.61) {};
\node[dot] (s4) at (6.30,1.55) {};
\draw[chunkarrow] (s0) to[out=16,in=190] (s1);
\draw[chunkarrow] (s1) to[out=12,in=186] (s2);
\draw[chunkarrow] (s2) to[out=6,in=180] (s3);
\draw[chunkarrow] (s3) to[out=-2,in=176] (s4);
\end{scope}
\node[note,anchor=south] (stateLabel) at (1.12,1.38) {$s_{t_k}$};
\path (s0) to[out=16,in=190] node[note,below=2pt] {$a_{t_k}$} (s1);

% The new plan begins only at the decision boundary.
\node[dot,fill=purple] (new1) at (7.58,.85) {};
\node[dot,fill=purple] (new2) at (8.84,.95) {};
\draw[chunkarrow,draw=purple] (s4) to[out=-20,in=175]
    node[note,above=2pt,text=purple] {} (new1);
\draw[chunkarrow,draw=purple] (new1) to[out=-5,in=190] (new2);
\node[note,text=purple,anchor=south] at (8.10,0.3) {$\tau_{k+1}$};
\node[dot,fill=loopGreen] (next1) at (10.12,1.67) {};
\node[dot,fill=loopGreen] (next2) at (11.40,1.52) {};
\node[dot,fill=loopGreen] (next3) at (12.68,1.70) {};
\draw[chunkarrow,draw=loopGreen] (new2) to[out=10,in=195] (next1);
\draw[chunkarrow,draw=loopGreen] (next1) to[out=-5,in=175] (next2);
\draw[chunkarrow,draw=loopGreen] (next2) to[out=5,in=190] (next3);
\node[note,text=loopGreen,anchor=west] at (12.80,1.70) {$\tau_{k+2}$};
% Robot poses at the commitment boundaries and an obstacle above the detour.
\pic[scale=.42,rotate=-20,opacity=.40] at (s4) {robot=loopInk};
\pic[scale=.42,rotate=10,opacity=.40] at (new2) {robot=loopInk};
\begin{scope}[xshift=.40cm,yshift=-.21cm]
\foreach \x/\y/\alpha in {9.14/1.40/.10,8.91/1.52/.18,8.68/1.65/.28,8.45/1.78/.40}
    \pic[scale=.42,rotate=150,opacity=\alpha] at (\x,\y) {robot=red!65!black};
\pic[scale=.42,rotate=150,opacity=.70] at (8.22,1.88) {robot=red!65!black};
\draw[red!65!black,opacity=.65,line width=.65pt,-{Stealth[length=3pt,width=2.5pt]}]
    (8.11,1.94) -- (7.79,2.12);
\node[note,text=red!65!black,anchor=south] at (9.65,1.95) {Moving obstacle};
\end{scope}
\end{scope}

% Equal control ticks; the selected budget determines the plan handoff.
% \draw[loopInk!45,line width=.5pt] (1.12,.57) -- (12.68,.57);
% \foreach \x in {1.12,2.42,3.72,5.02,6.30,7.58,8.84,10.12,11.40,12.68}
%     \draw[loopInk!45,line width=.5pt] (\x,.57) -- (\x,.70);
% \node[note,anchor=north] at (1.12,.49) {$t_k$};
% \node[note,anchor=north] at (2.90,.49) {$dt$ per step};
% \draw[purple!65,densely dashed,line width=.6pt] (6.30,1.48) -- (6.30,.57);
% \node[note,text=purple,anchor=north] at (6.30,.49) {$t_{k+1}=t_k+h_k$};
% \draw[loopGreen!65,densely dashed,line width=.6pt] (8.84,1.48) -- (8.84,.57);
% \node[note,text=loopGreen,anchor=north] at (9.70,.49) {$t_{k+2}=t_{k+1}+h_{k+1}$};
% \node[note,anchor=north] at (12.65,.49) {time};
\end{scope}
\end{tikzpicture}
\endgroup

%% file: chapters/related_work.tex
\subsection*{\textbf{Related Work}}
\label{subsec:related-work}
{We review the work most closely related to our setting.}
\paragraph{\textbf{Real-time and adaptive action chunking}}
Because a chunking policy emits several actions at once, a mechanism is required to cover the gap
while the next chunk is generated. RTC \cite{black2025rtc} keeps execution smooth
across that gap by conditioning the new chunk on the actions still running, and
DEFLECT \cite{zhu2026deflect} instead trains the policy to tolerate acting on
observations that have gone stale; both leave the chunk length fixed.
\cite{zhao2026dehp} makes that length adaptive, while \cite{chen2026dreamchunk} pre-computes several candidate
chunks and follows whichever best matches the realised state. Unlike our approach, none of these methods couples the execution length with the solver's computation budget, so a larger commitment cannot be leveraged to compute a better plan.

\paragraph{\textbf{Adaptive receding-horizon MPC}}
Event- and self-triggered MPC frameworks dynamically determine when to replan rather than solving at every control step, preserving feasibility and stability of the system \cite{sun2019robust}. Alternative approaches dictate the replanning interval based strictly on safety, defining it as the maximum duration the current trajectory can be executed while a collision-free fallback remains available \cite{magdici2017horizon}. However, these triggering mechanisms are designed fundamentally to \emph{conserve} computation subject to a baseline guarantee, treating the underlying optimizer as an oracle whose solution quality does not improve with additional allocated time. 

\paragraph{\textbf{Metareasoning and test-time computation}}
Metareasoning prices computation by the improvement in the decision quality it produces, net of the cost of running it~\cite{russell1991metareasoning,hay2012selecting}. This principle underlines anytime
algorithms, deliberation scheduling~\cite{dean1988timedependent,zilberstein1996anytime}, and learned computation budget~\cite{callaway2018bmps}. In these conventional formulations, the world is treated as frozen while the agent deliberates, so the cost of thinking is an explicit penalty chosen by the designer to trade against the expected benefit.

\paragraph{\textbf{Learning planning budgets}}
Closest to our work, \cite{muppidi2026budgets} learns how long to deliberate
in real time and casts the decision as a semi-MDP over budgeted options.
Their setting, however, is restricted to discrete game examples and therefore does not expose uncontrolled variation that arises when acting in the
real world. Furthermore, their method requires a separate reflex policy that acts in the open-loop phase, an assumption that our framework does not need. 

%% file: chapters/preliminaries.tex
% \vspace{-0.25cm}
\section{Preliminaries}
\label{sec:preliminaries}

Given a Markov Decision Process (MDP) defined by the tuple $\mathcal M:=(\mathcal S, \mathcal A, \mathcal P, r, \gamma)$, let $\mathcal{S} \subseteq \mathbb R^n$ be the state space, $\mathcal{A} \subseteq \mathbb R^m$ be the action space. We define $\mathcal P : \mathcal{S} \times \mathcal{A} \rightarrow \mathcal{S}$ as a transition kernel (dynamics model) yielding transition probabilities $\mathcal P(s_{t+1} \vert{} s_t, a_t)$. Moreover, let $r : \mathcal{S} \times \mathcal{A} \rightarrow \mathbb{R}$ and $g : \mathcal{S} \rightarrow \mathbb{R}$ be, respectively the one-step and the goal-conditioned reward functions. At any time $t$ the set of goals will be denoted by $G_t\subseteq S\times \mathcal A$. We work throughout in \emph{discrete time with continuous state}: where the index $t \in \mathbb Z_{\ge0}$ advances at the fixed control period $dt$. 

The objective of $\mathcal M$ is to determine a sequence of actions (control inputs), that drives the system into a desired state, solving the following infinite-horizon goal-conditioned control problem. 
\begin{equation}\customlabel{eq:control_problem}{$\mathrm{P_\mathcal{M}}$}
\mathrm{P_\mathcal{M}} : \max_{a_t} \sum_{t=0}^\infty \mathbb{E}\Big[\gamma^t \big( r(s_t, a_t) + g(s_t)\,\delta_{s_t\in G} \big)\Big]
\end{equation}
with  $\delta_{s_t\in G}$ is the indicator function equal to $1$ if the state $s_t$ has reached the goal (and $0$ otherwise), so that the terminal-style reward $g$ is collected only when the goal is reached.

To solve \ref{eq:control_problem} online in a receding horizon, we utilize a \textit{monotonic-improving} optimizer, $\psi$. A defining characteristic of such a solver is that its solution monotonically improves with the computational budget it is allocated. However, allocating more computation time introduces a temporal cost: the physical system must continue to operate while the solver thinks. To afford a computational budget of $h$ control periods, the solver must therefore commit to generating a sequence of actions (a chunk) of length $h \in \{1, \dots, \mathfrak h_{\mathrm{max}}\} = \mathcal{H}$, where $\mathfrak h_{\mathrm{max}} \in \mathbb N^+$. We define this solver as a mapping $\psi_h: \mathcal{S} \rightarrow \mathcal{T}_h$, where $\mathcal{T}_h = \mathcal{A}^h$ is the space of action sequences. By committing to this extended sequence, the robot executes the chunk open-loop, explicitly buying the required time window for the solver to concurrently compute the subsequent plan. While classic MPC formulations restrict this mapping to a single step ($h=1$), our formulation extends $h$ to leverage the solver's monotonic improvement at the expense of delayed reactivity. We evaluate this framework using Model Predictive Path Integral (MPPI) and the Cross Entropy Method (CEM), two standard sampling-based receding horizon solvers, though our method accommodates any monotonic-improving optimizer.

Two \textbf{fundamental difficulties} arise when solving \ref{eq:control_problem} online with a receding horizon time-monotone solver $\psi_h$ whose computation time is non-negligible and proportional to the action chunk length $h$:
\begin{mdframed}[
    linecolor=red!20,      % Sets the border color
    linewidth=0pt,        % Sets the border thickness
    roundcorner=5pt,        % Optional: rounds the corners slightly
    % hidealllines=true,
    backgroundcolor=red!5, 
    innertopmargin=5pt, 
    innerbottommargin=5pt
]
\emph{\textbf{(C1)} Open-loop execution creates vulnerability to unmodelled dynamics and disturbances.}
\end{mdframed}
 When the solver commits to an action chunk of length $h$, the system effectively operates without feedback for $h$ timesteps. During this interval, stochasticity, external disturbances, or discrepancies in the transition model $\mathcal{P}$ cause the true system state to diverge from the solver's predicted trajectory, leading to compounding errors or potential constraint violations.

\begin{mdframed}[
    linecolor=red!20,      % Sets the border color
    linewidth=0pt,        % Sets the border thickness
    roundcorner=5pt,        % Optional: rounds the corners slightly
    % hidealllines=true,
    backgroundcolor=red!5, 
    innertopmargin=5pt, 
    innerbottommargin=5pt
]
\emph{\textbf{(C2)} Finding the strict trade-off between plan optimality and system reactivity is a {non-trivial} problem.}
\end{mdframed}Because the solver's computational budget scales with the chunk length $h$, the problem formulation inherently couples how deeply the solver reasons with how frequently the system observes. A small $h$ enables rapid feedback to counter the disturbances described in (C1), but heavily restricts the computational budget, resulting in myopic and suboptimal planning. Conversely, a large $h$ allocates sufficient budget for deep, long-horizon deliberation but forces the system to remain open-loop for extended periods. A static $h$ therefore forces a rigid compromise between optimality and robustness, failing to exploit the fact that \textbf{different states in the MDP require different levels of deliberation versus reaction}.

We take up these challenges by adopting real-time chunking, which makes (C1) explicit rather than assuming it away, and by adapting the commitment length $h$ (and consequentially budget of the solver) online through an RL policy learnt from experience, which turns the design constant of (C2) into a decision. 

%% file: chapters/surprise_as_closed_loop_feedback.tex
\input{tikz/navigation_bars}
\input{tikz/navigation_ablation}
\section{Adaptive Computation Budgets as a semi-MDP}\label{subsec:smdp}

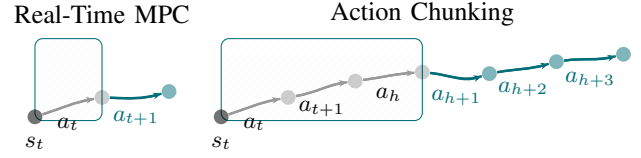
\begin{figure}[t]
    \centering
    \resizebox{\linewidth}{!}{\input{tikz/baselines}}
    \caption{\small {\textbf{Approaches}. (Left) Real-time
    MPC commits to one step ($h=1$), whereas action chunking (right) executes a fixed
    block of $h$ actions.}}
    \label{fig:epoch-clock}
    \vspace{-0.5cm}
\end{figure}

We formalise the problem as a semi-MDP (SMDP) \cite{sutton1999between}
$\mathfrak C = (\mathcal Z, \mathcal H, \mathcal M, \psi_h)$ over \emph{decision
epochs} $k = 0,1,2,\dots$, embedded in the underlying MDP $\mathcal M$ (whose environment time is $t$). At epoch $k$, as the computation budget scales with $h \in \mathcal{H}$ from Section~\ref{sec:preliminaries}, a {budget-}policy emits
\begin{equation}
  h_k \;\sim\; \Phi_\theta\!\left(\cdot \,\middle|\, z_k\right),
  \quad
  \Phi_\theta : \mathcal Z \to \Delta(\mathcal H),
  \label{eq:policy}
\end{equation}

where $\Delta(\mathcal H)$ is the simplex over $\mathcal H$. The output $h_k$ conditioned $\psi_{h_k}(\cdot | s_{t_k})$ on the underlying MDP $\mathcal M$ and fixes the next decision epoch at $t_{k+1} = t_k + h_k$ (see Fig.~\ref{fig:epoch-clock}).

During each epoch $k$, the agent accumulates the discounted step-level rewards of its committed chunk:

\begin{equation*}
R(s_{t_k}, h_k) = \sum_{j=0}^{h_k-1} \gamma^{\,j} \left[ r(s_{t_k+j}, a_{t_k+j}) + g(s_{t_k+j})\delta_{s_{t_k+j} \in G} \right].
\label{eq:smdp-reward}
\end{equation*}

By chaining these epoch rewards, we recover the original infinite-horizon objective \eqref{eq:control_problem}. The resulting semi-MDP objective, $\mathcal P_{\mathfrak C}$, mirrors the original control problem but discounts bootstrapping by the option's \emph{duration} instead of a single step:
\begin{equation}
  \mathcal P_{\mathfrak C}: \max_{\Phi_\theta}\;
  \mathbb E_{\mathcal P_{\mathcal M}^{\Phi_\theta,\psi_{h}}}\!\left[
    \sum_{k\ge 0} \gamma^{\,t_k}\, R(s_{t_k}, h_k)
  \right]
  \label{eq:smdp-objective}
\end{equation}
and $Q(s_{t_k},h_k) = R(s_{t_k},h_k) + \gamma^{\,h_k}\,\mathbb E\big[V(s_{t_{k+1}})\big]$, {where $\mathcal P_{\mathcal M}^{\Phi_\theta,\psi_h}$ denotes the transition kernel induced by the budget policy and planner on the underlying MDP $\mathcal M$.}

The construction is only meaningful if optimising \emph{when and how to think} optimises
the problem we started from, rather than a surrogate of it. Let $\rho(s,a) \coloneqq r(s,a) + g(s)\,\delta_{s\in G}$, $r$ and $g$ are bounded and $\gamma \in (0,1)$.

\begin{remark}[Equivalence of objectives]
\label{thm:equivalence}
Under the aforementioned assumptions, for every budget policy $\Phi_\theta$ and initial state $s_0$, the expected semi-MDP return equals the expected step-level return:
\begin{equation}
\small
\mathbb E_{\mathcal P_{\mathcal M}^{\Phi_\theta, \Psi}} \! \left[\sum_{k \ge 0} \gamma^{\,t_k} R(s_{t_k},h_k) \middle| s_0 \! \right] = \mathbb E_{\mathcal P} \! \left[\sum_{t \ge 0} \gamma^{\,t} \rho(s_t,a_t) \middle| s_0 \right]
\label{eq:equivalence}
\end{equation}
Consequently, maximizing the chunk-level semi-MDP objective $\mathcal{P}_{\mathfrak C}$ directly maximizes the underlying MDP objective $\mathcal{P}_{\mathcal M}$.
\end{remark}

\emph{Justification.}
This equivalence is a direct consequence of evaluating temporally extended actions within the Options framework \cite{sutton1999between}. By construction, the epoch intervals $I_k \coloneqq \{t_k, \dots, t_{k+1}-1\}$ form a strict, disjoint partition of the environment time ($\bigcup_{k \ge 0} I_k = \mathbb Z_{\ge 0}$). Because the intra-epoch return $R(s_{t_k}, h_k)$ is defined as the exact discounted sum of step-level rewards $\rho(s_t, a_t)$ over $I_k$, summing over all epochs pathwise seamlessly recovers the infinite-horizon return. Taking the expectation under the closed-loop trajectory distribution yields \eqref{eq:equivalence}.

\begin{remark}[relation to hierarchical RL] the construction is formally an
SMDP over options \cite{sutton1999between}, but the high-level action is a
\emph{duration}, not a behaviour: every budget executes the same chunk from the
same fixed $\psi$, so there is no option library to discover and no low-level
non-stationarity.
\end{remark}
\textbf{The physical value of computation is an action value.} The semi-MDP does not only state the problem: it collapses the informal notion
of physical value of computation as value inherited from the physics of the world,  into a one-line definition. As an ''action" (in the SMDP sense) is a budget $h$, the action-value function represents how much a stretch of deliberation is worth: $\Phi^{\mathrm{VoC}}_{\mathrm{abs}}:=Q^{\Phi}(s_{t_k}, h)$. From here,  we define the relative value of computation as the difference: 
\begin{equation}
  \Phi^{\mathrm{VoC}}_{\mathrm{rel}}(s_{t_k}, h_i, h_j)
  \coloneqq
  Q^{\Phi}(s_{t_k}, h_i) -  Q^{\Phi}(s_{t_k}, h_{j}),
  \label{eq:Phi-voc}
\end{equation}
with $ h_i, h_j\in \mathcal H$. Therefore, the optimal budget is defined as $h_k^\star = \arg\max_{h} \Phi^{\mathrm{VoC}}_{rel}(s_{t_k}, h, 1)$, which
 $\Phi_\theta$ \eqref{eq:policy} learns from input signal $z_k$. Optimizing $\Phi^{\mathrm{VoC}}_{rel}(s_{t_k}, h, 1)$, we aim at the optimal relative return gain from the trivial solution $h=1$.
 
\section{Bounded Divergence under Adaptive Commitment}

A natural question is how far open-loop execution can drift from the solver's prediction as the commitment grows.
For general solvers $\psi_h$, we bound the return deviation caused by open-loop execution under model mismatch, under the following assumption.
 
\begin{assumption}
\label{as:lipschitz}
The step-level reward $\rho(s,a) \coloneqq r(s,a) + g(s)\delta_{s \in G}$ is $L_\rho$-Lipschitz continuous with respect to the state. Furthermore, the true transition kernel $P(\cdot | s, a)$ and the solver's internal nominal model $\hat{P}(\cdot | s, a)$ exhibit a bounded single-step mismatch in the 1-Wasserstein distance, $W_1\big(P(\cdot | s, a), \hat{P}(\cdot | s, a)\big) \le \delta$, and the nominal dynamics are $L_P$-Lipschitz continuous in the state space under $W_1$.
\end{assumption}
 
\begin{theorem}[Bounded Open-Loop Deviation]
\label{thm:robust-margin}
Under Assumption~\ref{as:lipschitz}, let $e_0$ be the initial prediction error defined below. For any acceptable degradation $\epsilon \ge L_\rho e_0$, there exists a maximum open-loop commitment horizon $\overline{\mathfrak h}(\epsilon, e_0) \in \mathbb{N}_{\ge 1}\cup\{\infty\}$ such that executing any action chunk of length $h \le \overline{\mathfrak h}(\epsilon, e_0)$ keeps the expected cumulative open-loop return deviation bounded by $\epsilon$:
\begin{equation}
\mathbb{E} \left[ \sum_{j=0}^{h-1} \gamma^{\,j} \big| \rho(s_{t_k+j}, a_{t_k+j}) - \rho(\hat{s}_{t_k+j}, a_{t_k+j}) \big| \right] \le \epsilon.
\label{eq:optimality-bound}
\end{equation}
\end{theorem}
 
\begin{proof}
Let $e_j \coloneqq \mathbb{E}\big[\| s_{t_k+j} - \hat{s}_{t_k+j} \|\big]$ denote the expected state divergence at step $j$ of the chunk. Because the solver computes the current plan $\tau_k$ during the execution of the previous budget $h_{k-1}$, the new chunk begins from a predicted state $\hat{s}_{t_k}$ and inherits a non-zero initial divergence $e_0$.

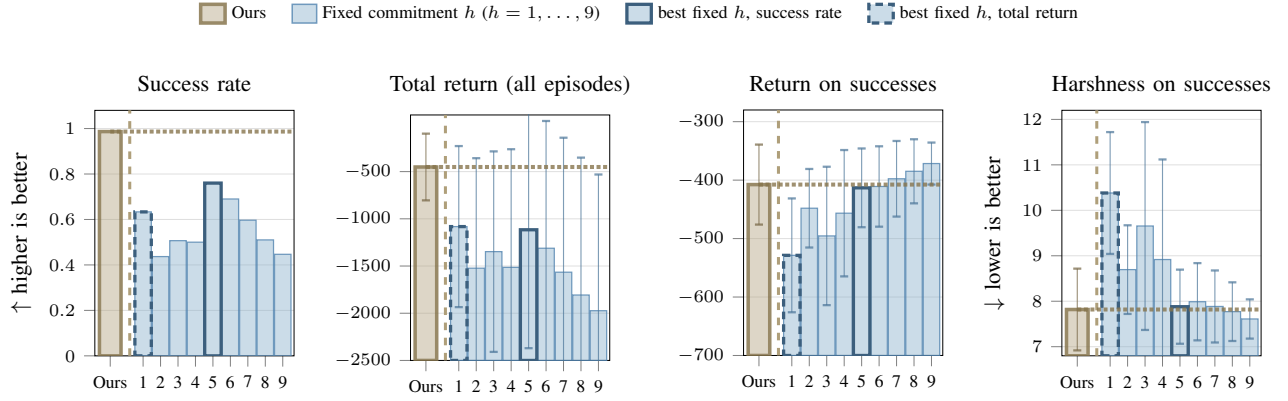
\begin{figure*}[t]
    \centering
    \resizebox{0.95\textwidth}{!}{%
    \begin{minipage}{\textwidth}
    \centering
    \ref*{navlegend}\\[2pt]
    \begin{tikzpicture}[baseline=(current bounding box.north)]
      \def\navhlsolid{\coordindex==4}\def\navhldash{\coordindex==0}%
      \navbars{title={Success rate}, noerrorbars, width=0.245\textwidth,
               ymin=0, ymax=1.08, ytick={0,0.2,0.4,0.6,0.8,1.0},
               ylabel={$\uparrow$ higher is better},
               legend to name=navlegend, legend columns=4,
               legend style={draw=none, fill=none, font=\scriptsize, text=black,
                             /tikz/every even column/.append style={column sep=8pt}},
               legend entries={Ours, {Fixed commitment $h$ ($h=1,\dots,9$)}, {best fixed $h$, success rate}, {best fixed $h$, total return}}}%
              {success}{0}
    \end{tikzpicture}\hfill
    \begin{tikzpicture}[baseline=(current bounding box.north)]
      \def\navhlsolid{\coordindex==4}\def\navhldash{\coordindex==0}%
      \navbars{title={Total return (all episodes)}, width=0.245\textwidth,
               ymin=0, ymax=2600, ytick={0,500,1000,1500,2000},
               yticklabels={$-2500$,$-2000$,$-1500$,$-1000$,$-500$}}%
              {total_reward}{-2500}
    \end{tikzpicture}\hfill
    \begin{tikzpicture}[baseline=(current bounding box.north)]
      \def\navhlsolid{\coordindex==4}\def\navhldash{\coordindex==0}%
      \navbars{title={Return on successes}, width=0.245\textwidth,
               ymin=0, ymax=420, ytick={0,100,200,300,400},
               yticklabels={$-700$,$-600$,$-500$,$-400$,$-300$}}%
              {reward}{-700}
    \end{tikzpicture}\hfill
    \begin{tikzpicture}[baseline=(current bounding box.north)]
      \def\navhlsolid{\coordindex==4}\def\navhldash{\coordindex==0}%
      \navbars{title={Harshness on successes}, width=0.245\textwidth,
               ymin=0, ymax=5.4, ytick={0.2,1.2,2.2,3.2,4.2,5.2},
               yticklabels={$7$,$8$,$9$,$10$,$11$,$12$},
               ylabel={$\downarrow$ lower is better}}%
              {harshness}{6.8}
    \end{tikzpicture}
    \end{minipage}%
    }
    \caption{\small \textbf{Navigation Results}, pooled over the three scenarios
    ($300$ randomised episodes). Our
    learned budget policy (left of the dashed rule) against every fixed
    commitment $h$ (blue).
    The two right-hand panels condition on successful episodes and therefore
    flatter the low-success baselines. The solid and dashed outlines mark the
    best fixed budget under success rate ($h=5$) and under total return
    ($h=1$): no single fixed $h$ is best at both, while the learned budget
    dominates both.}
    \label{fig:navigation-results}
\end{figure*}

By Assumption~\ref{as:lipschitz} and the Kantorovich--Rubinstein duality, the expected error propagates recursively as $e_{j} \le L_P e_{j-1} + \delta$. Unrolling this yields $e_j \le L_P^j e_0 + \delta \sum_{i=0}^{j-1} L_P^{\,i}$. The left-hand side of~\eqref{eq:optimality-bound} is therefore bounded by
\begin{equation*}
B(h) \coloneqq L_\rho \sum_{j=0}^{h-1} \gamma^{\,j} \Big( L_P^j e_0 + \delta \sum_{i=0}^{j-1} L_P^{\,i} \Big).
\end{equation*}
$B(1) = L_\rho e_0$, and $B(h)$ is nondecreasing in $h$ because it is a sum of nonnegative terms. Hence, for $\epsilon \ge L_\rho e_0$, the set $\{h : B(h) \le \epsilon\}$ is a nonempty initial segment of $\mathbb{N}_{\ge 1}$. Its supremum $\overline{\mathfrak h}(\epsilon, e_0)$ satisfies the claim.
\end{proof}
 
Theorem~\ref{thm:robust-margin} informs the design of the action space $\mathcal{H} = \{1, \dots, \mathfrak h_{\max}\}$. The lower bound $h=1$ gives the most frequent feedback and is admissible whenever the initial projection error $L_\rho e_0$ is acceptable. The upper bound should respect both the mismatch tolerance and the inherited initial divergence, $\mathfrak h_{\max} \le \overline{\mathfrak h}(\epsilon, e_0)$. 

\section{Optimization and Training}

The policy $\Phi_\theta$ conditions on a context $z_k$ that is deliberately richer than the
environment state, because the quantity being predicted (how much deliberation the current situation affords) depends on the plan in flight and on the solver's own confidence, not only on where the agent is:
\begin{equation}
  z_k=\big(\, s_{t_k},\ h_{k-1},\ \tau_k,\ \mathfrak b_k \,\big) \in \mathcal S\times \mathcal H\times\mathcal T^{H(h_{k-1})}\times\mathbb R^l
  \label{eq:context}
\end{equation}
The \textit{observation} $s_{t_k}$ is the state at the invocation,  $h_{k-1}$ is the \textit{previous budget};
the \textit{chunk to be executed} $\tau_k$, flattened over its steps, is the plan the agent will run open-loop \emph{while $\psi_h$ thinks}.

{\paragraph*{\textbf{Solver statistics}}
The last entry $\mathfrak b_k \in \mathbb R^l$ is a fixed-dimensional,
budget-invariant summary of the search that produced $\tau_k$. It is computed from
the rollout bundle and the latest execution feedback, giving the policy information
about planning difficulty without exposing a variable-length set of trajectories or
requiring an additional planner call. With $w_i$ denoting the softmax weight of
sampled rollout $i$ among $K$ candidates, the \emph{effective sample size}
$\mathrm{ESS} = \big(K\sum_i w_i^2\big)^{-1} \in (0,1]$ measures how concentrated
the search is: low values indicate
that a few rollouts dominate the update, whereas high values indicate
that many rollouts contribute comparably, as in a flat cost landscape
or a search that has already contracted around a solution. The
\emph{time-to-infeasibility} is the earliest predicted step at which a rollout
leaves the feasible set $\mathcal C$ and therefore measures the plan's remaining
safety margin. Finally, the \emph{dynamics-model mismatch} is the mean displacement
between predicted and realised positions of the objects intersected by the rollout
tube, providing an online measure of model reliability.}
$\mathfrak b_k$ is the term that distinguishes this formulation from a
state-conditioned schedule: it lets the policy condition on \emph{how hard the problem currently is for the solver}, which is observable to the planner but not to the state.

\subsection{Optimisation}
\label{subsec:optimisation}

We optimise $\Phi_\theta$ with PPO over decision epochs. Each transition in the
buffer is one epoch $\big(z_k,\, h_k,\, R(s_{t_k},h_k),\, \tau_k,\, z_{k+1}\big)$,
with $R$ accumulated as in \eqref{eq:smdp-reward} and advantages estimated by a
duration-aware GAE,
\begin{equation*}
  \delta_k = R_k + \gamma^{\,h_k}\, V(z_{k+1}) - V(z_k),
  \quad
  A_k = \delta_k + \gamma^{\,h_k}\,\lambda\, A_{k+1},
  \label{eq:gae}
\end{equation*}
so that both the bootstrap and the eligibility decay follow elapsed environment
time rather than epoch count. The actor and critic share an encoder over
$z_k$ and the actor emits a categorical distribution over $\mathcal H$.

\begin{proposition}[Monotonic Improvement]
Assuming exact advantages and exact
maximisation of the PPO surrogate at each iteration, updates of $\Phi_\theta$
monotonically improve $\mathcal P_{\mathfrak C}$ and, by Remark~1, $\mathrm P_{\mathcal M}$.
\end{proposition}
\begin{proof}
Let $\tilde P(z'|z,h)\coloneqq\gamma^{h-1}P(z'|z,h)$, with the remaining mass
$1-\gamma^{h-1}$ sent to an absorbing zero-reward state. Then $\mathfrak C$ is a
standard $\gamma$-discounted MDP with identical values, since
$\gamma\,\gamma^{h-1}=\gamma^{h}$. PPO is an instance of mirror
learning~\cite{kuba2022mirror}, whose updates monotonically improve such MDPs;
Remark~1 transfers the improvement to $\mathrm P_{\mathcal M}$.
\end{proof}

%% file: tikz/navigation_bars.tex
% Navigation results as bars: Ours | dashed rule | fixed taus, almost touching.
% Data: figures/navigation/data/nav_{rl,fixed}.dat -- swap the CSVs, nothing else.
% Ours sits at x = 1.00 (width 1.05); fixed bars at x = 0.85*tau + 1.75,
% i.e. 2.60 ... 10.25 with width 0.8, so a 0.05 gap between them.

\definecolor{rlall}{HTML}{AF9F7A}
\definecolor{rlallE}{HTML}{87754D}
\definecolor{rlnop}{HTML}{B079A9}
\definecolor{rlnopE}{HTML}{8A4789}
\definecolor{rlplo}{HTML}{85ADC9}
\definecolor{rlploE}{HTML}{5587B2}

\def\navhlsolid{0}
\def\navhldash{0}

\pgfplotsset{
  navaxis/.style={
    width=0.345\textwidth, height=5.0cm,
    ybar, bar width=1.05,
    /pgf/bar shift=0pt,
    xmin=0.25, xmax=9.95,
    xtick={1.00,2.60,3.45,4.30,5.15,6.00,6.85,7.70,8.55,9.40},
    xticklabels={Ours,1,2,3,4,5,6,7,8,9},
    xticklabel style={font=\scriptsize},
    yticklabel style={font=\scriptsize},
    label style={font=\small},
    title style={font=\small, yshift=-2pt},
    tick align=outside, tick pos=left,
    ymajorgrids, grid style={gray!25, line width=0.3pt},
    error bars/error mark=|,
  },
  % a panel whose spread is not meaningful drops the whiskers entirely
  noerrorbars/.style={every axis plot post/.append style={error bars/y dir=none}},
  barstyle/.style 2 args={fill=#1, draw=#2, line width=0.5pt,
                          fill opacity=0.42, draw opacity=0.85,
                          error bars/y dir=both, error bars/y explicit,
                          error bars/error bar style={#2!85!black,
                            line width=0.45pt, draw opacity=0.7},
                          % caps drawn heavier than the whisker itself
                          error bars/error mark options={#2!85!black,
                            line width=0.9pt, mark size=1.6pt, draw opacity=0.7},
                          area legend,
                          legend image code/.code={%
                            \filldraw[fill=#1, draw=#2, line width=0.6pt,
                                      fill opacity=0.42]
                              (0cm,-0.13cm) rectangle (0.34cm,0.13cm);}},
  pickrow/.style={y filter/.expression={\coordindex==#1 ? y : nan}},
  % outlines drawn over chosen fixed baselines; see \navhlsolid / \navhldash
  highlight/.style={draw=rlploE!70!black, fill=none, line width=1.3pt,
                    draw opacity=1, bar width=0.8, error bars/y dir=none,
                    area legend,
                    legend image code/.code={%
                      \filldraw[draw=rlploE!70!black, line width=1.3pt,
                                fill=rlplo, fill opacity=0.42]
                        (0cm,-0.13cm) rectangle (0.34cm,0.13cm);}},
  highlightd/.style={highlight, densely dashed,
                    legend image code/.code={%
                      \filldraw[draw=rlploE!70!black, line width=1.3pt,
                                densely dashed, fill=rlplo, fill opacity=0.42]
                        (0cm,-0.13cm) rectangle (0.34cm,0.13cm);}},
}

% #1 axis options  #2 CSV column  #3 baseline (bars grow from here)
\newcommand{\navbars}[3]{%
  % level reached by Ours, for the dotted comparison rule
  \pgfplotstableread[col sep=comma]{figures/navigation/data/nav_rl.dat}\navrltable
  \pgfplotstablegetelem{0}{#2}\of\navrltable
  \pgfmathsetmacro{\oursy}{\pgfplotsretval-(#3)}%
  \begin{axis}[navaxis, #1]
    \addplot[barstyle={rlall}{rlallE}, pickrow=0, line width=1.4pt,
             legend image code/.code={%
               \filldraw[fill=rlall, draw=rlallE, line width=1.4pt,
                         fill opacity=0.42]
                 (0cm,-0.13cm) rectangle (0.34cm,0.13cm);}]
      table[col sep=comma, x expr={1.00},
            y expr={\thisrow{#2}-(#3)}, y error=#2_std]
      {figures/navigation/data/nav_rl.dat};
    \addplot[barstyle={rlplo}{rlploE}, bar width=0.8,
             y filter/.expression={\thisrow{tau}<10 ? y : nan}]
      table[col sep=comma, x expr={0.85*\thisrow{tau}+1.75},
            y expr={\thisrow{#2}-(#3)}, y error=#2_std]
      {figures/navigation/data/nav_fixed.dat};
    \draw[dashed, rlall, line width=1.2pt]
      ({axis cs:1.95,0}|-{rel axis cs:0,0}) -- ({axis cs:1.95,0}|-{rel axis cs:0,1});
    \addplot[highlight,
             y filter/.expression={\thisrow{tau}<10 ? ((\navhlsolid) ? y : nan) : nan}]
      table[col sep=comma, x expr={0.85*\thisrow{tau}+1.75},
            y expr={\thisrow{#2}-(#3)}]
      {figures/navigation/data/nav_fixed.dat};
    \addplot[highlightd,
             y filter/.expression={\thisrow{tau}<10 ? ((\navhldash) ? y : nan) : nan}]
      table[col sep=comma, x expr={0.85*\thisrow{tau}+1.75},
            y expr={\thisrow{#2}-(#3)}]
      {figures/navigation/data/nav_fixed.dat};
    \draw[densely dotted, rlallE, opacity=0.8, line width=1.6pt]
      ({axis cs:1.00,\oursy}) -- ({rel axis cs:1,0}|-{axis cs:0,\oursy});
  \end{axis}%
}

%% file: tikz/navigation_ablation.tex
% Ablation bars: full | full minus planner statistics | only planner statistics.
% Data: figures/navigation/data/ablation.dat (pooled 300 episodes per variant),
% regenerated by figures/navigation/data/build.py. Bar styles come from
% tikz/navigation_bars, which must be \input before this file.

\definecolor{ablteal}{HTML}{00665E}
\definecolor{ablblue}{HTML}{247BB5}
\definecolor{ablpurple}{HTML}{512579}

\pgfplotsset{
  ablaxis/.style={
    width=0.21\textwidth, height=3.8cm,
    ybar, bar width=0.62, /pgf/bar shift=0pt,
    xmin=0.4, xmax=3.6, xtick={1,2,3},
    xticklabels={Full, {Full$-$PS}, {PS only}},
    xticklabel style={font=\scriptsize},
    yticklabel style={font=\scriptsize},
    label style={font=\small},
    title style={font=\small, yshift=-2pt},
    tick align=outside, tick pos=left,
    ymajorgrids, grid style={gray!25, line width=0.3pt},
    error bars/error mark=|,
  },
}

% #1 axis options  #2 CSV column  #3 baseline (bars grow from here)
\newcommand{\ablbars}[3]{%
  \begin{axis}[ablaxis, #1]
    \addplot[barstyle={ablteal}{ablteal}, fill opacity=0.6, draw opacity=1, line width=1.2pt,
      mark=*, mark size=1.1pt, mark options={draw=ablteal!70!black, fill=ablteal!70!black,
        fill opacity=1, draw opacity=1}, pickrow=0] table[col sep=comma,
      x expr={\thisrow{id}==0 ? 1 : 4-\thisrow{id}}, y expr={\thisrow{#2}-(#3)}, y error=#2_std]
      {figures/navigation/data/ablation.dat};
    \addplot[barstyle={ablblue}{ablblue}, fill opacity=0.6, draw opacity=1, line width=1.2pt,
      mark=*, mark size=1.1pt, mark options={draw=ablblue!70!black, fill=ablblue!70!black,
        fill opacity=1, draw opacity=1}, pickrow=1] table[col sep=comma,
      x expr={\thisrow{id}==0 ? 1 : 4-\thisrow{id}}, y expr={\thisrow{#2}-(#3)}, y error=#2_std]
      {figures/navigation/data/ablation.dat};
    \addplot[barstyle={ablpurple}{ablpurple}, fill opacity=0.6, draw opacity=1, line width=1.2pt,
      mark=*, mark size=1.1pt, mark options={draw=ablpurple!70!black, fill=ablpurple!70!black,
        fill opacity=1, draw opacity=1}, pickrow=2] table[col sep=comma,
      x expr={\thisrow{id}==0 ? 1 : 4-\thisrow{id}}, y expr={\thisrow{#2}-(#3)}, y error=#2_std]
      {figures/navigation/data/ablation.dat};
  \end{axis}%
}

%% file: tikz/baselines.tex
% ------------------------------------------------------------------
% Requires:  \usepackage{tikz}
%            \usetikzlibrary{arrows.meta,positioning,backgrounds}
%            (backgrounds is what provides the "background" pgf layer)
% ------------------------------------------------------------------
% Two baselines stacked, same drawing, only the commitment changes:
%   TOP   -- Real-Time MPC:   h = 1
%   BOTTOM -- Action Chunking: h = 3 (fixed)
% ------------------------------------------------------------------
\definecolor{darkteal}{RGB}{0,110,120}
\colorlet{chunkcol}{black}
\begin{tikzpicture}[
    >=Stealth,
    dot/.style={circle, fill=chunkcol, inner sep=2.0pt, fill opacity=0.50},
    traj/.style={line width=0.9pt, draw=chunkcol},
    chunkarrow/.style={-{Stealth[length=3.2pt, width=2.6pt]}, line width=0.9pt, draw=chunkcol},
    alabel/.style={font=\small,text opacity=1},
    title/.style={font=\normalsize},
]

%% ==================================================================
%% TOP -- Real-Time MPC: the robot commits to a single action, h = 1
%% ==================================================================

\node[dot,fill opacity=.55] (s) at (0, 0) {};
\begin{scope}[opacity=.40,transparency group]

\node[dot] (a1) at (0.9, 0.26) {};

\draw[traj] (s) to[out=18, in=190] (a1);
\draw[chunkarrow] (s) to[out=18, in=190] (a1);
\end{scope}
\node[alabel,below=1pt of s] {$s_t$};
\path (s) to[out=18,in=190] node[alabel,below=1pt] {$a_t$} (a1);
\node[title] at (.90,1.38) {Real-Time MPC};

\begin{pgfonlayer}{background}
    \draw[draw=darkteal, rounded corners=3pt]
        (0, -0.05) rectangle (0.9, 1.05);
    % faint teal hatching over the grey fill
    \draw[draw=none, rounded corners=3pt, pattern=north east lines,
          pattern color=darkteal, opacity=0.18]
        (0, -0.05) rectangle (0.9, 1.05);
\end{pgfonlayer}

% --- the new chunk tau' takes over right after the single committed step ---
\begin{scope}
\colorlet{chunkcol}{darkteal}
% only the first action of tau' is actually executed: keep it solid
\node[dot] (n1) at (1.8, 0.34) {};
\draw[traj] (a1) to[out=-2, in=196] (n1);
\draw[chunkarrow] (a1) to[out=-2, in=196]
    node[alabel,below=1pt,text=darkteal] {$a_{t+1}$} (n1);
\end{scope}

%% ==================================================================
%% BOTTOM -- Action Chunking: h > 1 is fixed (here h = 3)
%% ==================================================================

\begin{scope}[xshift=2.5cm]

  \node[dot,fill opacity=.55] (s2) at (0, 0) {};
  \begin{scope}[opacity=.40,transparency group]

  \node[dot] (b1) at (0.9, 0.26) {};
  \node[dot] (b2) at (1.8, 0.48) {};
  \node[dot] (b3) at (2.7, 0.60) {};

  \draw[traj]
      (s2) to[out=18, in=190]
      (b1) to[out=10, in=188]
      (b2) to[out=8, in=184]
      (b3);

  \draw[chunkarrow] (s2) to[out=18, in=190] (b1);
  \draw[chunkarrow] (b1) to[out=10, in=188] (b2);
  \draw[chunkarrow] (b2) to[out=8, in=184] (b3);
  \end{scope}
  \node[alabel,below=1pt of s2] {$s_t$};
  \path (s2) to[out=18,in=190] node[alabel,below=1pt] {$a_t$} (b1);
  \path (b1) to[out=10,in=188] node[alabel,below=1pt] {$a_{t+1}$} (b2);
  \path (b2) to[out=8,in=184] node[alabel,below=1pt] {$a_h$} (b3);
  \node[title] at (2.70,1.38) {Action Chunking};

  \begin{pgfonlayer}{background}
      \draw[draw=darkteal, rounded corners=3pt]
          (0, -0.05) rectangle (2.7, 1.05);
      % faint teal hatching over the grey fill
      \draw[draw=none, rounded corners=3pt, pattern=north east lines,
            pattern color=darkteal, opacity=0.18]
          (0, -0.05) rectangle (2.7, 1.05);
  \end{pgfonlayer}

  % --- the new chunk tau' only takes over after the whole chunk is executed ---
  \begin{scope}
  \colorlet{chunkcol}{darkteal}
  % the whole chunk is committed and executed: tau' stays opaque
  \node[dot] (r0) at (3.6, 0.58) {};
  \node[dot] (r1) at (4.5, 0.74) {};
  \node[dot] (r2) at (5.4, 0.84) {};
  \draw[traj] (b3) to[out=-6, in=198] (r0) to[out=10, in=196] (r1)
                   to[out=8, in=194] (r2);
  \draw[chunkarrow] (b3) to[out=-6, in=198]
      node[alabel,below=1pt,text=darkteal] {$a_{h+1}$} (r0);
  \draw[chunkarrow] (r0) to[out=10, in=196]
      node[alabel,below=1pt,text=darkteal] {$a_{h+2}$} (r1);
  \draw[chunkarrow] (r1) to[out=8, in=194]
      node[alabel,below=1pt,text=darkteal] {$a_{h+3}$} (r2);
  \end{scope}

\end{scope}

\end{tikzpicture}

%% file: chapters/experiments.tex
\section{Experiments}
\label{sec:experiments}
\begin{figure*}[t]
    \centering
    \setlength{\tabcolsep}{1pt}\renewcommand{\arraystretch}{0}%
    \begin{tabular}{@{}cccc@{}}
      \includegraphics[width=0.233\textwidth,trim=9bp 0 9bp 0,clip]{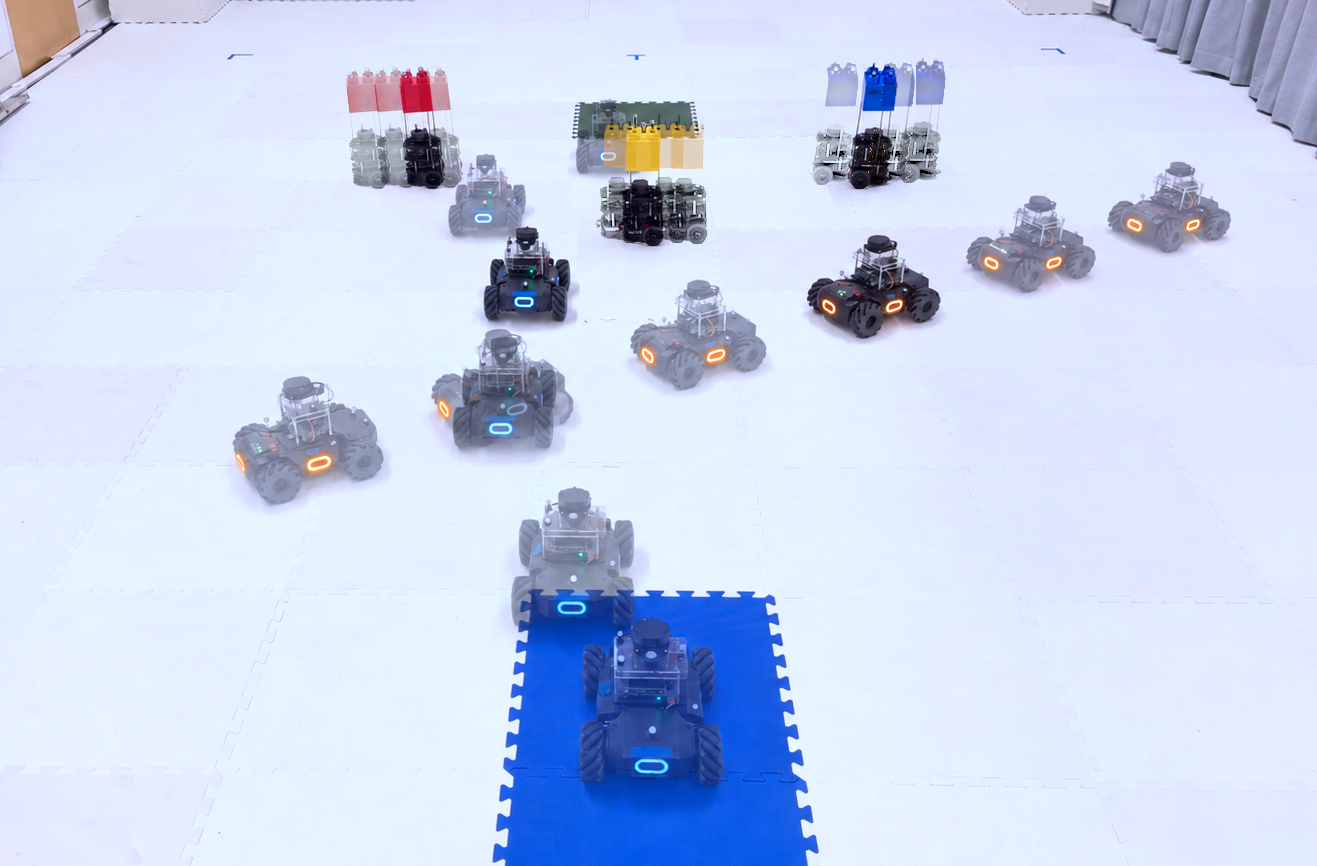} &
      \includegraphics[width=0.233\textwidth,trim=0 3.17bp 0 3.17bp,clip]{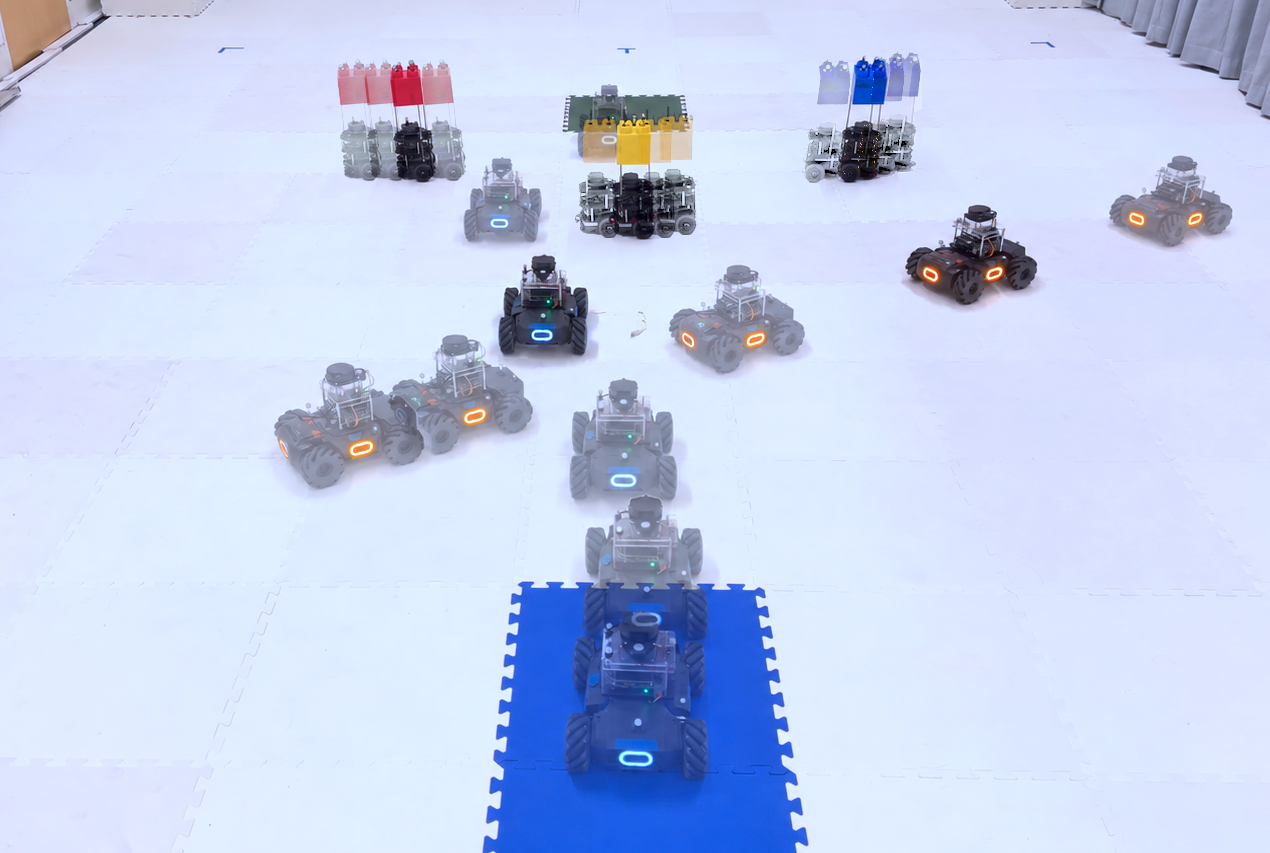} &
      \includegraphics[width=0.233\textwidth,trim=43.25bp 0 43.25bp 0,clip]{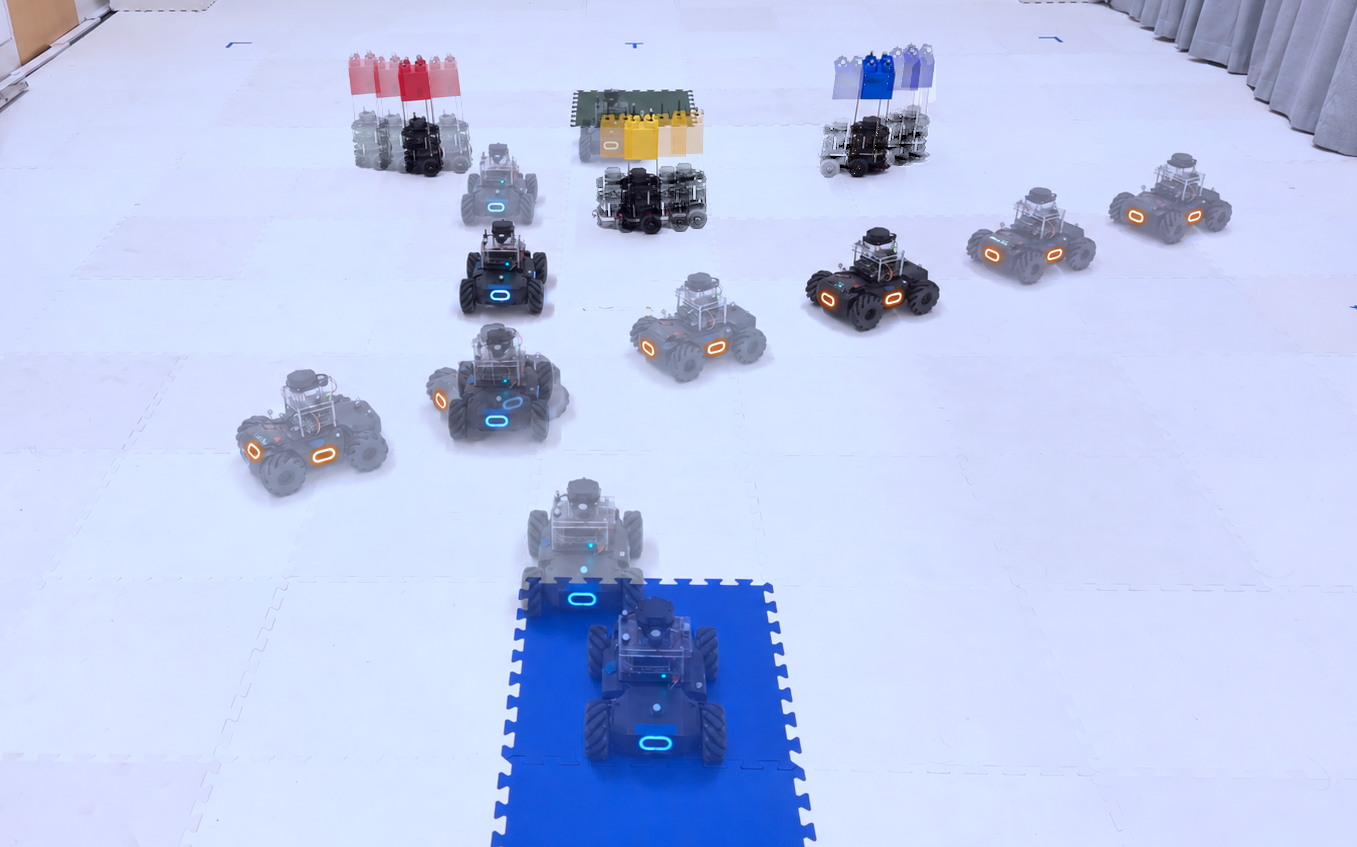} &
      \includegraphics[width=0.233\textwidth,trim=38.75bp 0 38.75bp 0,clip]{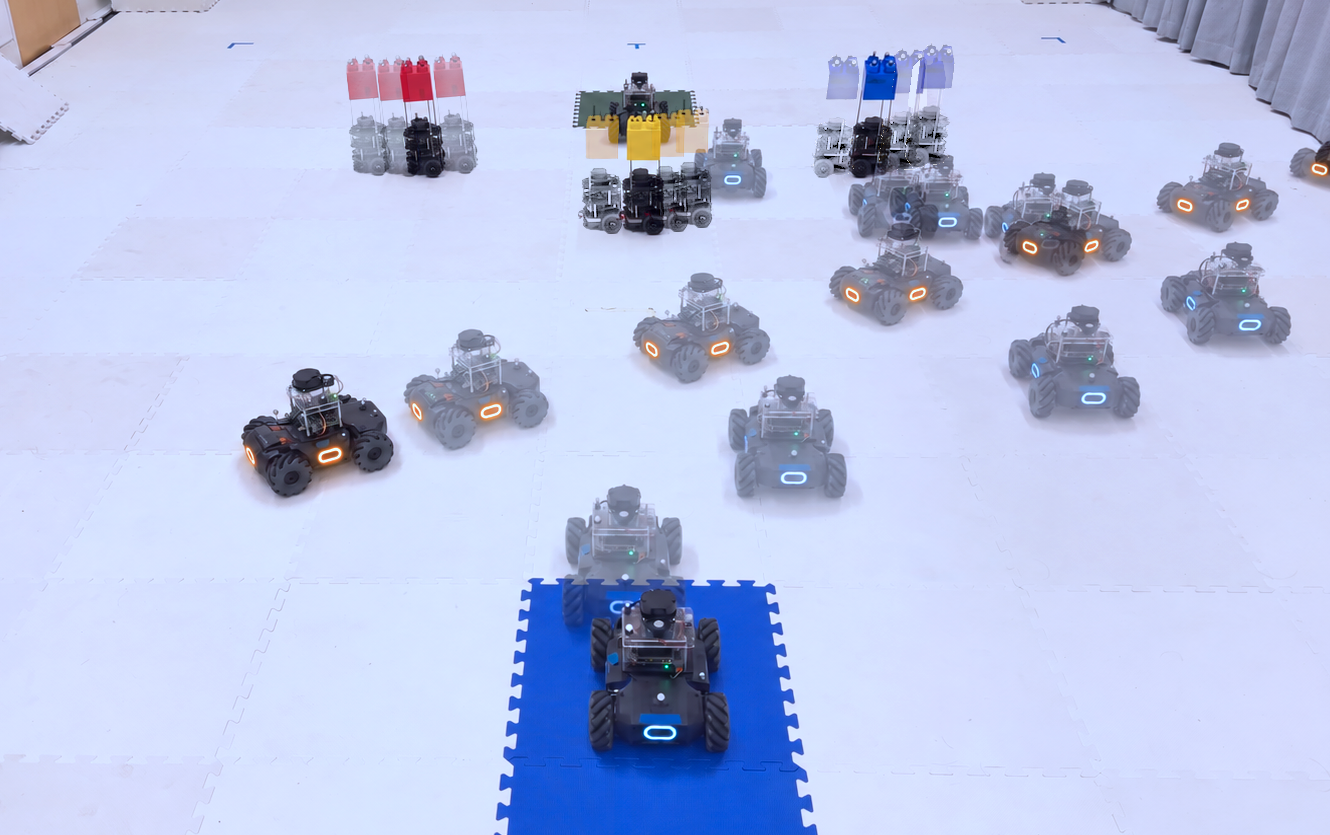}\\[2pt]
      {\scriptsize Ours} & {\scriptsize $h=9$} & {\scriptsize $h=4$}
      & {\scriptsize $h=1$}
    \end{tabular}\\[4pt]
    \input{tikz/navigation_real_arena}
    \vspace{-0.5cm}
    \caption{\small \textbf{Real-world Merging Lane scenario.} \emph{Top:}
    temporal composites on the RoboMaster platform under our learned budget and
    fixed commitments $h\in\{1,4,9\}$; each panel overlays successive robot
    poses from one episode. \emph{Bottom:} the commitment our policy selects
    along its episode, against the three fixed budgets. The budget collapses to
    $h=1$ while the robot merges past the crossing obstacle (red band) and
    while it threads the close passage between the static obstacles. 
    }
    \label{fig:navigation-real}
\end{figure*}

Our experimental evaluation is designed to validate the efficacy of adaptive computation budgets across diverse robotic domains. Specifically, we seek to answer the following research questions:
\begin{itemize}
    \item \textbf{Q1} {\textit{(Performance)}:} Does dynamic adaptation of the computational budget and execution horizon consistently outperform fixed-budget baselines in dynamic and constrained environments?
    \item \textbf{Q2} {\textit{(Adaptation Mechanism)}:} Can the learned policy correctly identify state-dependent requirements for deliberation versus reactivity, and how critical are internal planner statistics to this arbitration?
    \item \textbf{Q3} {\textit{(Boundary Behavior)}:} How does the adaptive policy behave in non-smooth cost landscapes where the optimal commitment horizon lacks strong state-dependent variation?
\end{itemize}

To quantitatively evaluate these questions, we report two primary metrics:
\textbf{(1) task success rate}, the fraction of evaluation episodes that satisfy
the task-specific goal criterion in goal-conditioned scenarios; and
\textbf{(2) mean total return}, the undiscounted sum of environment rewards
over an episode, averaged over all evaluation episodes, including failures.
Higher values are better for both metrics.

For navigation tasks, we additionally measure \textbf{harshness}, a
dimensionless measure combining path tortuosity and motion jerk,
$H=(L/D)[1+\log(1+J_{\mathrm{norm}})]$.
Here, $L$ is the travelled path length, and $D$ is the straight-line distance
between the robot's initial and final positions. The normalized jerk
$J_{\mathrm{norm}}$ measures how abruptly acceleration changes, using the
accumulated squared jerk normalized by episode duration and path length.
Lower harshness indicates a
more direct and smoother trajectory.

We also present the \textbf{optimality regret of a fixed-commitment baseline}
at each decision boundary along its trajectory as the relative value of computation with respect to the optimal budget-action $h^*$ at state $s_t$:
\begin{equation}
    \Phi^{\mathrm{VoC}}_{\mathrm{rel}}(s_{t_k}, h^*, h)=\Phi^{\mathrm{VoC}}_{abs}(s_t, h^*) - Q^{\Phi}(s_t,h).
  \label{eq:exp-suboptimality-gap} 
\end{equation}
Thus, $ \Phi^{\mathrm{VoC}}_{\mathrm{rel}}(s_{t_k}, h^*, h)$ measures the estimated return lost by choosing the fixed
commitment instead of the best available commitment under
$\Phi_{\theta}(s_t)$ (lower values are better).

We deploy our framework across three distinct scenarios, systematically chosen to stress the computation-reactivity trade-off from different physical dimensions: holonomic ground navigation, deployed in the real-world; quadrotor with suspended cable payload; and contact-rich planar manipulation (3D Push-T). The properties of each system are described in the related paragraphs.

\textbf{Implementation Details.}
The hardware and software infrastructure is standardized across experiments to ensure consistent solver performance. The underlying control and communication stack is orchestrated via ROS~2. Hardware experiments are conducted on a DJI RoboMaster platform \cite{blumenkamp2024cambridge}  for the holonomic ground navigation. Simulated environments utilize MuJoCo \cite{todorov2012mujoco} for the quadrotor dynamics and the Push-T library using a \cite{kurtz2024hydrax} for planar manipulation. To achieve real-time, high-frequency planning, the predictive model and the Model Predictive Path Integral (MPPI) solver are implemented entirely in JAX and executed via MuJoCo MJX \cite{todorov2012mujoco}. All neural network inference and hardware-accelerated planning are executed on a single NVIDIA A100 GPU (40GB VRAM), ensuring the solver's wall-clock latency strictly dictates the physical temporal cost of computation.
{The control period is $dt=0.1\,\mathrm{s}$ for navigation, $dt=0.02\,\mathrm{s}$ for cable-payload flight, and $dt=0.04\,\mathrm{s}$ for PushT; a commitment of $h$ therefore corresponds to $h\,dt$ seconds of open-loop execution.}
{The budget policy is a two-layer tanh MLP (128 units per layer), with a categorical actor and scalar critic. PPO uses $\lambda=0.95$, learning rate $3\times10^{-4}$, minibatches of 256, four epochs, clip $0.2$, and gradient clipping at $0.5$. We train for 2,500 updates with 2,048 transitions per update.}
{In the experiments, MPPI scales its prediction horizon linearly with $h$ and scales its samples linearly for navigation and as $\sqrt{h}$ for cable flight; CEM on PushT scales both horizon and population linearly.}

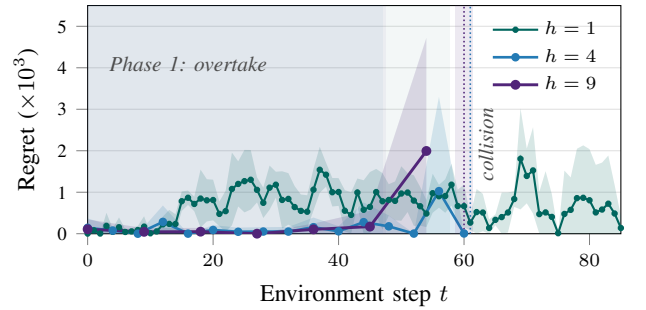
\begin{figure}[!t]
  \centering
  \input{tikz/suboptimality_gap}
  \caption{\small \textbf{Optimality Regret} in the Merging Lane scenario along three fixed commitment trajectories.
  Lines are means and standard deviations over $10$ episodes. The tint marks Phase~1, before the moving obstacle has been passed, which ends at a different step for each commitment. Dotted rules mark the mean collision step for each fixed strategy.}
  \label{fig:suboptimality-gap}
  \vspace{-0.3cm}
\end{figure}

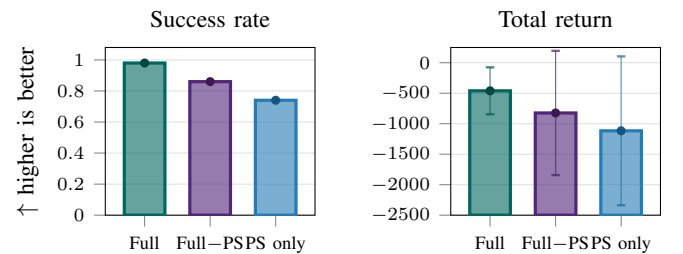
\begin{figure}[!t]
    \centering
    \begin{tikzpicture}
      \ablbars{title={Success rate}, noerrorbars, width=0.245\textwidth,
               ymin=0, ymax=1.08, ytick={0,0.2,0.4,0.6,0.8,1.0},
               ylabel={$\uparrow$ higher is better}}%
              {success}{0}
    \end{tikzpicture}\hfill
    \begin{tikzpicture}
      \ablbars{title={Total return}, width=0.245\textwidth,
               ymin=0, ymax=2750, ytick={0,500,1000,1500,2000,2500},
               yticklabels={$-2500$,$-2000$,$-1500$,$-1000$,$-500$,$0$}}%
              {total_reward}{-2500}
    \end{tikzpicture}
    \caption{\small \textbf{Holonomic 2D navigation: planner statistics ablation}. Ablation of the policy input, pooled over the three
    scenarios ($300$ episodes per variant). \textsc{Full} uses
    $(s_{t_k},h_{k-1},\tau_k,\mathfrak b_k)$; \textsc{PS only} uses the planner
    statistics $\mathfrak b_k$ alone; \textsc{Full$-$PS} uses everything except
    $\mathfrak b_k$. The full input achieves the highest success rate and
    total return, while removing planner statistics drops success by $12$
    percentage points.}
    \label{fig:navigation-ablation}
    \vspace{-0.3cm}
\end{figure}

\begin{figure*}[!t]
  \centering
  \setlength{\tabcolsep}{1pt}\renewcommand{\arraystretch}{0}%
  \begin{tabular}{@{}cccc@{}}
    \includegraphics[width=0.247\textwidth,trim=47.51bp 0 47.51bp 74.72bp,clip]{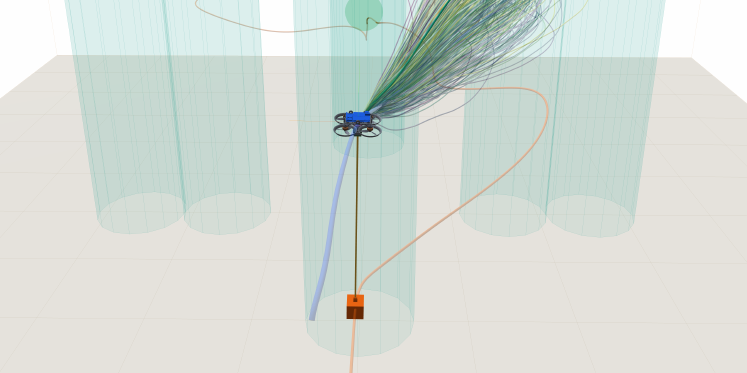} &
    \includegraphics[width=0.247\textwidth,trim=47.51bp 20bp 47.51bp 54.72bp,clip]{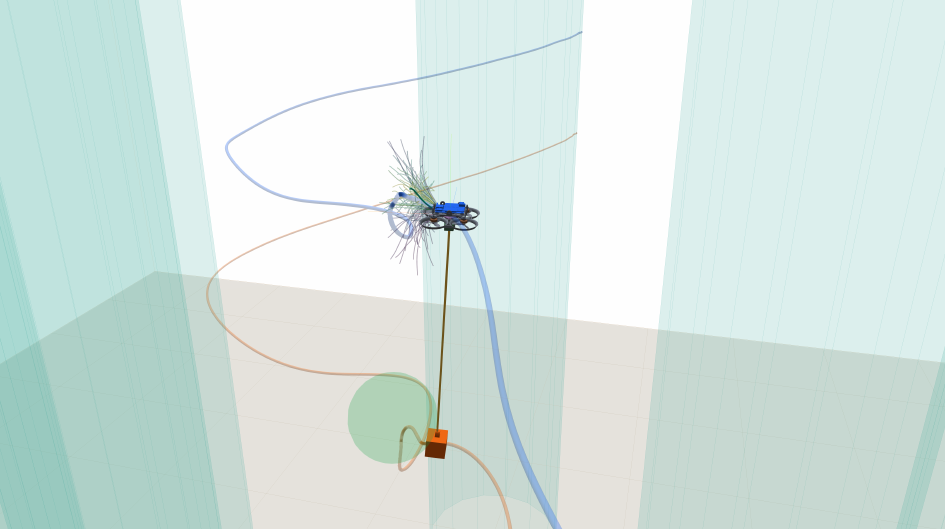} &
    \includegraphics[width=0.247\textwidth,trim=47.51bp 10bp 47.51bp 64.72bp,clip]{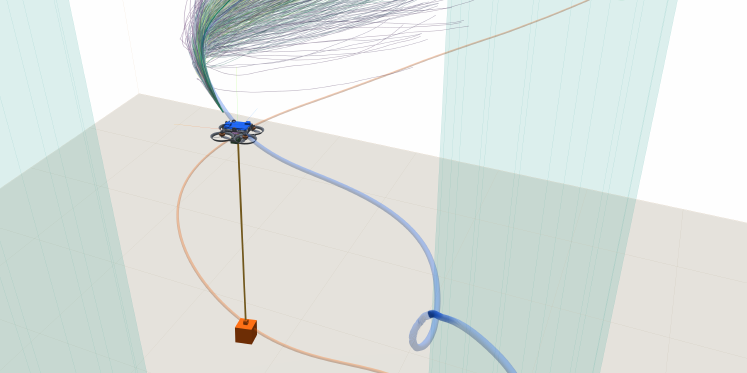} &
    \includegraphics[width=0.247\textwidth,trim=47.51bp 9.51bp 47.51bp 37.89bp,clip]{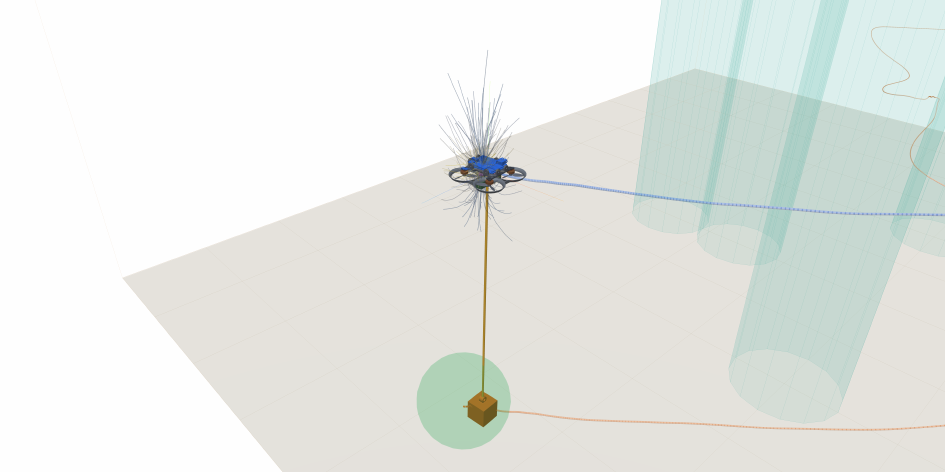} \\[2pt]
    {\scriptsize (a) Approach first target, $h=8$} &
    {\scriptsize (b) Stabilise, $h=2$} &
    {\scriptsize (c) Unlock second target, $h=8$} &
    {\scriptsize (d) Approach final target, $h=3$}
  \end{tabular}
  \caption{\small \textbf{Dynamic commitment during cable-payload flight.}
  Four stages from one episode show the learned policy increasing the
  commitment for transit and reducing it near each target.}
  \label{fig:cable-payload-dynamic-h}
  \vspace{-0.3cm}
\end{figure*}
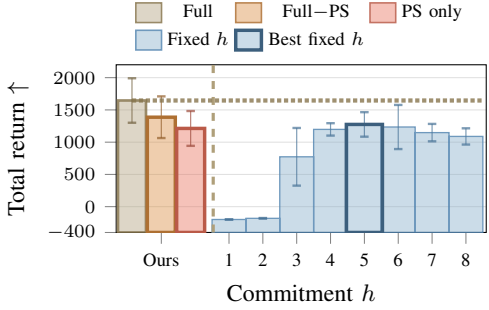
\begin{figure}[!t]
  \centering
  \input{tikz/cable_return}
  \caption{\small \textbf{Cable-payload slalom return.} Mean total return over
  $20$ episodes per method. The left group compares the full policy (gold) with
  ablations without planner statistics (orange) and using planner statistics only
  (salmon); fixed commitments are shown in blue.}
  \label{fig:cable-return}
  \vspace{-0.38cm}
\end{figure}
\subsection{Holonomic 2D navigation}
\label{subsec:exp-nav}
In this navigation example, because the dynamics are simple, failures arise strictly from failing to balance the deep search required for global optimality against the rapid feedback needed to avoid dynamic obstacles.
{We consider three scenarios with distinct obstacle geometries and dynamics: a fixed wall with three stochastic obstacles (``\textit{barrier}'', Fig.~\ref{fig:hero}, right), one crossing robot with three stochastic obstacles (``\textit{merging lane}'', Fig.~\ref{fig:navigation-real}), and five laterally moving obstacles (``\textit{highway}'', Fig.~\ref{fig:hero}, centre). The environment reward is
$\rho=-\lVert p_t-p_g\rVert_2^2-3000\,\bm{1}_{\mathrm{collision}}$,
where $p_t$ and $p_g$ are the robot and goal positions. Success or collision terminates the episode.}
{Aggregate and per-scenario results are shown in Fig.~\ref{fig:navigation-results} and Table~\ref{tab:navigation-per-scenario}. The adaptive policy achieves the strongest overall performance across success rate, total return, and harshness.}
{In the \textit{merging lane} scenario (Fig.~\ref{fig:navigation-real}), the policy uses a larger budget early in the episode to identify a safe route, then reduces the commitment to increase feedback while passing the crossing robot and nearby stochastic obstacles. Around $t=3\,\mathrm{s}$, it approaches the reactive limit $h_{\min}=1$.}
Figure~\ref{fig:suboptimality-gap} plots the
optimality regret, $\Phi_{\mathrm{rel}}^{\mathrm{VOC}}, $~\eqref{eq:exp-suboptimality-gap} of fixed commitment $h$
at each decision boundary. 

\textbf{Sim to Real}. The RoboMaster evaluation uses the same learned policy and budget-to-planner mapping as simulation, with only the physical robot, sensing, and ROS~2 execution stack changed. The close agreement with the simulated navigation behaviour indicates a small simulation-to-real gap for this task.

\textbf{\emph{\checkmark \, Planner statistics guide adaptation.}}
The solver summary $\mathfrak b_k$ is what tells the policy when to spend and
when to react. {The ablation in Fig.~\ref{fig:navigation-ablation} shows that
removing this feedback lowers success from $98\%$ to $86\%$; feedback alone reaches $74\%$ success, showing that state
and solver signals are complementary.}

\begin{table}[t]
  \centering\footnotesize
  \setlength{\tabcolsep}{3.5pt}
  \caption{\small \textbf{Navigation results} ($100$ episodes per scenario). We compare our policy with the same three fixed budgets selected from the pooled results in Fig.~\ref{fig:navigation-results}: $h=5$ (best Success rate), $h=1$ (best total Return), and $h=9$ (best Harshness). The budgets remain fixed across scenarios; total return is averaged over all episodes.}
  \label{tab:navigation-per-scenario}
  \begin{tabular}{llccc}
    \toprule
    Scenario & Strategy & Success $\uparrow$ & Total return $\uparrow$ & Harshness $\downarrow$\\
    \midrule
    \multirow{4}{*}{\shortstack[l]{Barrier\\[1pt]}}
      & \cellcolor{rlall!30}\textbf{Ours} & \cellcolor{rlall!30}$\boldsymbol{0.97}$ & \cellcolor{rlall!30}$\boldsymbol{-435}$ & \cellcolor{rlall!30}${7.37}$ \\
      & $h=5$ (S) & $0.73$ & $-1167$ & $\bm{7.28}$\\
      & $h=1$ (R) & $0.12$ & $-1523$ & $7.42$ \\
      & $h=9$ (H) & $0.63$ & $-1400$ & $7.55$ \\
    \midrule
    \multirow{4}{*}{\shortstack[l]{Merging Lane\\[1pt]}}
      & \cellcolor{rlall!30}\textbf{Ours} & \cellcolor{rlall!30}$\boldsymbol{1.00}$ & \cellcolor{rlall!30}$\boldsymbol{-385}$ & \cellcolor{rlall!30}$\boldsymbol{7.02}$\\
      & $h=5$ (S) & $0.63$ & $-1462$ & $7.17$ \\
      & $h=1$ (R) & $\bm{1.00}$ & $-460$ & $9.68$ \\
      & $h=9$ (H) & $0.61$ & $-1466$ & $7.46$ \\
    \midrule
    \multirow{4}{*}{\shortstack[l]{Highway\\[1pt]}}
      & \cellcolor{rlall!30}\textbf{Ours} & \cellcolor{rlall!30}$\boldsymbol{0.99}$ & \cellcolor{rlall!30}$\boldsymbol{-534}$ & \cellcolor{rlall!30}${9.05}$ \\
      & $h=5$ (S) & $0.92$ & $-719$ & $\bm{8.84}$ \\
      & $h=1$ (R) & $0.78$ & $-1265$ & $11.73$\\
      & $h=9$ (H) & $0.08$ & $-2988$ & $9.26$ \\
    \bottomrule
  \end{tabular}
  \vspace{-0.3cm}
\end{table}

\subsection{Quadrotor with a suspended cable payload}
\label{subsec:exp-drone}

The quadrotor carries a cable-suspended load through a two-goal slalom. The
stiff, underactuated dynamics punish staleness via open-loop system divergence
rather than environmental changes. This requires the system to balance deep,
computationally expensive sampling for stable transit against highly reactive
short-horizon commitments near targets. Its dense reward is
$r_t=r_t^{\mathrm{goal}}+r_t^{\mathrm{velocity}}
+r_t^{\mathrm{hover}}+r_t^{\mathrm{smooth}}+r_t^{\mathrm{swing}}
+r_t^{\mathrm{collision}}$. For distance $d_t$ and braking-aware target velocity
$v_t^*$, $r_t^{\mathrm{goal}}=\exp(-2d_t)$ and
$r_t^{\mathrm{velocity}}=2\exp(-\lVert v_t-v_t^*\rVert_2^2/0.35^2)$.
The remaining terms reward Gaussian hover accuracy, penalize action/body rates,
and set $r_t^{\mathrm{swing}}=-0.5\lVert\omega_t^{\mathrm{cable}}\rVert_2^2$ and
$r_t^{\mathrm{collision}}=-300(\bm{1}_{\mathrm{tip}}+
\bm {1}_{\mathrm{drone}})$; intermediate settling adds $250$ plus up to
$100$ for precision. {Here $d_t$ is the distance to the active target, $v_t$ is the quadrotor translational velocity, $v_t^*$ is the braking-aware target velocity, and $\omega_t^{\mathrm{cable}}$ is the payload's cable angular velocity.} 
A robot trial is successful only when the system reaches the first goal, remains within the goal and stability tolerances for the required settling window, and then reaches and stabilises at the final goal under the same criterion. Our policy succeeds in $80\%$ of trials, while every fixed
commitment fails, {and it discovers a two-stage computation pattern: it uses larger budgets during stable transit and shorter commitments near the targets, achieving the highest return among the fixed strategies (Fig.~\ref{fig:cable-return}).} Furthermore, Fig.~\ref{fig:cable-return} confirms that \textit{planner} \textit{statistics} help the policy achieving the best performance.

\subsection{Contact-rich planar manipulation}
\label{subsec:exp-pusht}

\begin{figure}[!t]
  \centering
  \begin{minipage}[c]{0.42\linewidth}
    \includegraphics[width=\linewidth,trim=49.878bp 0 49.878bp 0,clip]{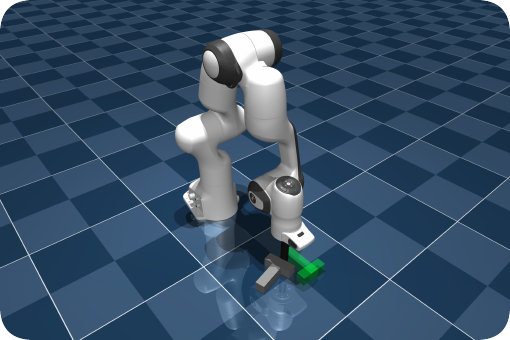}
  \end{minipage}\hfill%
  \begin{minipage}[c]{0.56\linewidth}
  \centering\scriptsize
  \setlength{\tabcolsep}{2pt}
  \begin{tabular}{@{}ccc@{}}
    \toprule
    $h$ & $\Delta x$ [cm] $\downarrow$ & $\Delta\theta$ [rad] $\downarrow$\\
    \midrule
    $1$ & $9.41\pm1.55$ & $0.090\pm0.055$\\
    $2$ & $8.22\pm2.65$ & $0.100\pm0.066$\\
    $3$ & $5.79\pm4.50$ & $0.049\pm0.057$\\
    $4$ & $4.19\pm4.79$ & $0.044\pm0.056$\\
    $5$ & $4.48\pm4.64$ & $0.054\pm0.060$\\
    $6$ & $4.95\pm4.64$ & $0.072\pm0.071$\\
    $7$ & $1.21\pm1.65$ & $0.018\pm0.023$\\
    $8$ & $\boldsymbol{0.65}\pm0.48$ & $0.020\pm0.017$\\
    $9$ & $1.13\pm1.46$ & $0.022\pm0.018$\\
    $10$ & $0.76\pm0.71$ & $\boldsymbol{0.016}\pm0.014$\\
    \midrule
    \rowcolor{rlall!30}
    \textbf{Ours} & $\boldsymbol{0.65}\pm0.48$ & $0.020\pm0.017$\\
    \bottomrule
  \end{tabular}
  \end{minipage}
  \caption{\small \textbf{Boundary behaviour on the 3D PushT environment}. Left: robot-arm manipulation setup. Right: terminal position and orientation errors after $600$ steps, reported as mean $\pm$ sample standard deviation over $16$ matched validation episodes. Our policy selects $\bar h=8.00$.}
  \label{fig:pusht-results}
  \vspace{-0.2cm}
\end{figure}

In Push-T~\cite{chi2023diffusion}, a planar pusher slides a free T-shaped block
onto a target using the hydrax model~\cite{kurtz2024hydrax} and CEM in MuJoCo
MJX. With position error $d_t$, angular error $\delta_t$, pusher--block distance
$q_t$, and normalized action $a_t$, its environment reward is
$r_t=-\min\{10,\,0.01[(d_t/0.02)^2+(\delta_t/0.15)^2]
+2q_t^2+0.001\lVert a_t\rVert_2^2\}$. The budget policy receives the decrease
in the bracketed pose cost across its chunk. Changing contacts makes plans
stale and tests local budget sensitivity; results are in
Fig.~\ref{fig:pusht-results}.

{It can be seen that the policy collapses to $h=8$ everywhere, matching the best fixed baseline. We attribute this to the nominal setting: planning and execution share exact dynamics, the system is fully observed and disturbance-free, and the objective does not reward computational savings or earlier arrival, leaving little incentive for frequent feedback.}

%% file: tikz/navigation_real_arena.tex
% Landscape top-down view of the four real-world Merging Lane episodes (top,
% start at the left edge, goal at the right) and, below it, the commitment our
% policy selected along its own episode.
% Data: figures/navigation/data/real/ (build.py); the CSVs are already rotated,
% column x is the world y and column y is the world x, hence y dir=reverse.
% The arena is stretched along the travel direction, so the scene is drawn with
% fixed-size marks rather than data-space circles.
\begin{tikzpicture}
\definecolor{salmon}{HTML}{E8705A}
\definecolor{hotpink}{HTML}{E0218A}
\pgfplotsset{
  realpath/.style={line width=1.6pt, no marks, opacity=0.5},
  realdec/.style={only marks, mark=*, mark size=1.0pt, draw opacity=0,
                  fill opacity=0.75},
  realblob/.style 2 args={only marks, mark=*, mark size=#1, fill=#2,
                          fill opacity=0.35, draw=#2!60!black, draw opacity=0.7,
                          line width=0.3pt},
}

\begin{axis}[name=tau,
  width=0.77\textwidth, height=0.115\textwidth, scale only axis,
  xmin=0, xmax=7.2, ymin=0, ymax=13.5,
  ytick={0,2,4,6,8}, xtick={0,2,4,6,7.2},
  tick label style={font=\scriptsize}, label style={font=\footnotesize},
  xlabel={time [s]}, xlabel style={yshift=2pt}, ylabel={$h\sim\Phi_{\theta}$ [steps]},
  tick align=outside, tick pos=left,
  axis background/.style={fill=white},
  grid=both, grid style={black!10, line width=0.3pt},
  legend style={font=\scriptsize, at={(1.03,0.5)}, anchor=west,
                draw=none, fill=none, inner sep=1pt},
]
  % the two phases the episode passes through
  \addplot[forget plot, draw=none, fill=orange, fill opacity=0.12]
    coordinates {(0,0) (1.7,0) (1.7,13.5) (0,13.5)} -- cycle;
  \addplot[forget plot, draw=none, fill=red, fill opacity=0.12]
    coordinates {(1.7,0) (2.4,0) (2.4,13.5) (1.7,13.5)} -- cycle;
  \addplot[forget plot, draw=none, fill=hotpink, fill opacity=0.12]
    coordinates {(3.8,0) (6.5,0) (6.5,13.5) (3.8,13.5)} -- cycle;
  \node[font=\scriptsize, orange!60!black, anchor=west]
    at (axis cs:0.08,11.9) {Finding optimal path};
  \node[font=\scriptsize, red!60!black, anchor=center]
    at (axis cs:2.05,9.9) {overtake};
  \node[font=\scriptsize, hotpink!70!black, anchor=center] at (axis cs:5.15,11.9)
    {close passage, noisy obstacles};
  % the fixed budgets, for reference
  \addplot[forget plot, gaprose, densely dotted, line width=0.7pt, no marks,
    opacity=0.7] coordinates {(0,1) (10,1)};
  \addplot[forget plot, gappurple, densely dotted, line width=0.7pt, no marks,
    opacity=0.7] coordinates {(0,4) (10,4)};
  \addplot[forget plot, gapblue, densely dotted, line width=0.7pt, no marks,
    opacity=0.7] coordinates {(0,9) (10,9)};
  \addplot[salmon, line width=1.1pt, no marks, opacity=0.9, forget plot]
    table[col sep=comma, x=t, y=tau]{figures/navigation/data/real/ours_tau.dat};
  % legend key: the line with one decision dot on it
  \addlegendimage{salmon, line width=1.1pt, opacity=0.9, mark=*, mark size=1.0pt,
    mark options={fill=salmon, draw=salmon}}
  \addlegendentry{\shortstack[l]{dynamic chunking\\achieved by\\our method}}
  % one translucent dot per decision, i.e. where a new commitment is chosen
  \addplot[only marks, mark=*, mark size=1.0pt, salmon,
    forget plot] table[col sep=comma, x=t, y=tau]
    {figures/navigation/data/real/ours_tau_dec.dat};
\end{axis}
\end{tikzpicture}

%% file: tikz/suboptimality_gap.tex
% First panel of navigation_steps_overlay_dynamic_obstacle_three_static.pdf.
% Mean +/- one sample std over 5 episodes (seeds 10000-10004); dotted rules are
% mean collision steps, the strip around one is its std. Data and provenance:
% figures/navigation/data/suboptimality/ (build.py rebuilds the h*.dat files).
\begin{tikzpicture}
\definecolor{gapteal}{HTML}{00665E}
\definecolor{gapblue}{HTML}{247BB5}
\definecolor{gappurple}{HTML}{512579}
\pgfplotsset{
  gapband/.style={draw=none, fill=#1, fill opacity=0.16, forget plot},
  gapphase/.style={draw=none, fill=#1, fill opacity=0.05, forget plot},
  gapcoll/.style={#1, densely dotted, line width=0.6pt, forget plot},
}
\begin{axis}[
  width=\columnwidth, height=4.6cm,
  xlabel={Environment step $t$}, ylabel={Regret ($\times 10^3$)},
  xmin=0, xmax=85, ymin=0, ymax=5500,
  xtick={0,20,40,60,80}, ytick={0,1000,2000,3000,4000,5000},
  yticklabels={0,1,2,3,4,5}, scaled y ticks=false,
  tick label style={font=\scriptsize}, label style={font=\small},
  tick align=outside, tick pos=left, ymajorgrids, grid style={black!12},
  legend style={at={(0.98,0.98)}, anchor=north east,
    draw=none, fill=none, font=\scriptsize},
]
% phase 1 ends when each fixed commitment has passed the moving obstacle
\addplot[gapphase=gapteal]   coordinates {(0,0) (57.8,0) (57.8,5500) (0,5500)} -- cycle;
\addplot[gapphase=gapblue]   coordinates {(0,0) (47.0,0) (47.0,5500) (0,5500)} -- cycle;
\addplot[gapphase=gappurple] coordinates {(0,0) (47.6,0) (47.6,5500) (0,5500)} -- cycle;
\node[font=\footnotesize\itshape, text=black!70, anchor=north west]
  at (axis cs:2,4500) {Phase 1: overtake};
% mean +/- std bands
\addplot[name path=h1lo, draw=none, forget plot]
  table[col sep=comma,x=step,y=lo]{figures/navigation/data/suboptimality/h1.dat};
\addplot[name path=h1hi, draw=none, forget plot]
  table[col sep=comma,x=step,y=hi]{figures/navigation/data/suboptimality/h1.dat};
\addplot[gapband=gapteal] fill between[of=h1lo and h1hi];
\addplot[name path=h4lo, draw=none, forget plot]
  table[col sep=comma,x=step,y=lo]{figures/navigation/data/suboptimality/h4.dat};
\addplot[name path=h4hi, draw=none, forget plot]
  table[col sep=comma,x=step,y=hi]{figures/navigation/data/suboptimality/h4.dat};
\addplot[gapband=gapblue] fill between[of=h4lo and h4hi];
\addplot[name path=h9lo, draw=none, forget plot]
  table[col sep=comma,x=step,y=lo]{figures/navigation/data/suboptimality/h9.dat};
\addplot[name path=h9hi, draw=none, forget plot]
  table[col sep=comma,x=step,y=hi]{figures/navigation/data/suboptimality/h9.dat};
\addplot[gapband=gappurple] fill between[of=h9lo and h9hi];
% means
\addplot[gapteal, line width=0.8pt, mark=*, mark size=0.8pt]
  table[col sep=comma,x=step,y=mean]{figures/navigation/data/suboptimality/h1.dat};
\addlegendentry{$h=1$}
\addplot[gapblue, line width=1pt, mark=*, mark size=1.1pt]
  table[col sep=comma,x=step,y=mean]{figures/navigation/data/suboptimality/h4.dat};
\addlegendentry{$h=4$}
\addplot[gappurple, line width=1.1pt, mark=*, mark size=1.3pt]
  table[col sep=comma,x=step,y=mean]{figures/navigation/data/suboptimality/h9.dat};
\addlegendentry{$h=9$}
% mean collision steps; the strip is one std around the h=9 mean
\addplot[gapcoll=gapblue] coordinates {(61,0) (61,5500)};
\addplot[draw=none, fill=gappurple, fill opacity=0.1, forget plot]
  coordinates {(58.59,0) (61.41,0) (61.41,5500) (58.59,5500)} -- cycle;
\addplot[gapcoll=gappurple] coordinates {(60,0) (60,5500)};
\node[font=\footnotesize\itshape, text=black!70, rotate=90, anchor=west]
  at (axis cs:63.5,1050) {collision};
\end{axis}
\end{tikzpicture}

%% file: tikz/cable_return.tex
% Reuse the navigation bar styles from Figure 4.
\definecolor{ablora}{HTML}{D98C3F}
\definecolor{ablsal}{HTML}{E8705A}
\begin{tikzpicture}
  \begin{axis}[navaxis,
    width=0.75\columnwidth, height=3.8cm,
    xmin=0.00, xmax=8.20,
    xtick={1.00,2.50,3.25,4.00,4.75,5.50,6.25,7.00,7.75},
    xticklabels={Ours,1,2,3,4,5,6,7,8},
    ymin=0, ymax=2600,
    ytick={0,400,900,1400,1900,2400},
    yticklabels={$-400$,$0$,$500$,$1000$,$1500$,$2000$},
    xlabel={Commitment $h$}, ylabel={Total return $\uparrow$},
    legend columns=3,
    legend style={draw=none, fill=none, font=\scriptsize,
                  at={(0.5,1.03)}, anchor=south},
  ]
    % Shift the bar baseline to -400, as in Figure 4's return panels.
    \addplot[barstyle={rlall}{rlallE}, bar width=0.62, line width=1.0pt,
             y filter/.expression={y+400}]
      coordinates {(0.35,1646) +- (0,345)};
    \addlegendentry{Full}
    \addplot[barstyle={ablora}{ablora!80!black}, bar width=0.62,
             y filter/.expression={y+400}]
      coordinates {(1.00,1387) +- (0,323)};
    \addlegendentry{Full$-$PS}
    \addplot[barstyle={ablsal}{ablsal!80!black}, bar width=0.62,
             y filter/.expression={y+400}]
      coordinates {(1.65,1213) +- (0,270)};
    \addlegendentry{PS only}
    \addplot[draw=ablora!80!black, fill=none, bar width=0.62,
             line width=1.3pt, error bars/y dir=none,
             y filter/.expression={y+400}, forget plot]
      coordinates {(1.00,1387)};
    \addplot[draw=ablsal!80!black, fill=none, bar width=0.62,
             line width=1.3pt, error bars/y dir=none,
             y filter/.expression={y+400}, forget plot]
      coordinates {(1.65,1213)};
    \addplot[barstyle={rlplo}{rlploE}, bar width=0.75,
             y filter/.expression={y+400}]
      coordinates {
        (2.50,-199) +- (0,4)
        (3.25,-181) +- (0,5)
        (4.00,773)  +- (0,447)
        (4.75,1198) +- (0,96)
        (5.50,1275) +- (0,190)
        (6.25,1235) +- (0,342)
        (7.00,1148) +- (0,135)
        (7.75,1089) +- (0,126)
      };
    \addlegendentry{Fixed $h$}
    \addplot[highlight, y filter/.expression={y+400}]
      coordinates {(5.50,1275)};
    \addlegendentry{Best fixed $h$}
    \draw[dashed, rlall, line width=1.2pt]
      (axis cs:2.15,0) -- (axis cs:2.15,2600);
    \draw[densely dotted, rlallE, opacity=0.8, line width=1.6pt]
      (axis cs:0.35,2046) -- (axis cs:8.10,2046);
  \end{axis}
\end{tikzpicture}

%% file: chapters/conclusion.tex
\vspace{0.3cm}
\section{Conclusion}
\label{sec:conclusion}

Robot \textit{metareasoning} is a fundamental capability of embodied
intelligence: an agent must decide how much to deliberate under
environmental uncertainty and temporal constraints. We introduce a semi-MDP framework for robot metareasoning that learns adaptive computation budgets for real-time control. By leveraging planner statistics, the policy dynamically balances plan execution length and planner computation time based on \textit{plan quality} and \textit{uncertainty}. 
Across three environments with distinct sources of difficulty, the policy adapts its budget to the demands of the task. 
Our results show that online adaptation is \textit{highly beneficial} when exploitable budget variation exists.
It discovers longer commitments for global route, smooth transit and shorter ones near dynamic obstacles, critical passage and, in cable-payload flight, near targets and disturbances. 
In contact-rich planar manipulation, the learned
policy converges to a constant budget matching the best fixed baseline, indicating that adaptation is unnecessary when the nominal task provides little exploitable variation.

\textbf{Future work} will explore richer, high-dimensional planner statistics, learned world models, and more expressive planning algorithms.

%% file: main.bbl
\begin{thebibliography}{10}
\providecommand{\url}[1]{#1}
\csname url@samestyle\endcsname
\providecommand{\newblock}{\relax}
\providecommand{\bibinfo}[2]{#2}
\providecommand{\BIBentrySTDinterwordspacing}{\spaceskip=0pt\relax}
\providecommand{\BIBentryALTinterwordstretchfactor}{4}
\providecommand{\BIBentryALTinterwordspacing}{\spaceskip=\fontdimen2\font plus
\BIBentryALTinterwordstretchfactor\fontdimen3\font minus
  \fontdimen4\font\relax}
\providecommand{\BIBforeignlanguage}[2]{{%
\expandafter\ifx\csname l@#1\endcsname\relax
\typeout{** WARNING: IEEEtran.bst: No hyphenation pattern has been}%
\typeout{** loaded for the language `#1'. Using the pattern for}%
\typeout{** the default language instead.}%
\else
\language=\csname l@#1\endcsname
\fi
#2}}
\providecommand{\BIBdecl}{\relax}
\BIBdecl

\bibitem{otto2019opportunity}
A.~R. Otto and N.~D. Daw, ``The opportunity cost of time modulates cognitive
  effort,'' \emph{Neuropsychologia}, vol. 123, pp. 92--105, 2019.

\bibitem{daw2005uncertainty}
N.~D. Daw, Y.~Niv, and P.~Dayan, ``Uncertainty-based competition between
  prefrontal and dorsolateral striatal systems for behavioral control,''
  \emph{Nature Neuroscience}, vol.~8, no.~12, pp. 1704--1711, 2005.

\bibitem{keramati2011speed}
M.~Keramati, A.~Dezfouli, and P.~Piray, ``Speed/accuracy trade-off between the
  habitual and the goal-directed processes,'' \emph{PLoS Computational
  Biology}, vol.~7, no.~5, p. e1002055, 2011.

\bibitem{shenhav2017toil}
A.~Shenhav, S.~Musslick, F.~Lieder, W.~Kool, T.~L. Griffiths, J.~D. Cohen, and
  M.~M. Botvinick, ``Toward a rational and mechanistic account of mental
  effort,'' \emph{Annual Review of Neuroscience}, vol.~40, pp. 99--124, 2017.

\bibitem{russell1991metareasoning}
S.~Russell and E.~Wefald, ``Principles of metareasoning,'' \emph{Artificial
  Intelligence}, vol.~49, no. 1--3, pp. 361--395, 1991.

\bibitem{hay2012selecting}
N.~Hay, S.~Russell, D.~Tolpin, and S.~E. Shimony, ``Selecting computations:
  Theory and applications,'' in \emph{Proc. Conf. Uncertainty in Artificial
  Intelligence (UAI)}, 2012.

\bibitem{williams2017mppi}
G.~Williams, N.~Wagener, B.~Goldfain, P.~Drews, J.~M. Rehg, B.~Boots, and E.~A.
  Theodorou, ``Information theoretic {MPC} for model-based reinforcement
  learning,'' in \emph{Proc. IEEE Int. Conf. Robotics and Automation (ICRA)},
  2017, pp. 1714--1721.

\bibitem{bharadhwaj2020cem}
H.~Bharadhwaj, K.~Xie, and F.~Shkurti, ``Model-predictive control via
  cross-entropy and gradient-based optimization,'' in \emph{Proceedings of the
  Conference on Learning for Dynamics and Control}, 2020, pp. 277--286.

\bibitem{zhao2023act}
T.~Z. Zhao, V.~Kumar, S.~Levine, and C.~Finn, ``Learning fine-grained bimanual
  manipulation with low-cost hardware,'' in \emph{Proc. Robotics: Science and
  Systems (RSS)}, 2023.

\bibitem{chi2023diffusion}
C.~Chi, S.~Feng, Y.~Du, Z.~Xu, E.~Cousineau, B.~Burchfiel, and S.~Song,
  ``Diffusion policy: Visuomotor policy learning via action diffusion,'' in
  \emph{Proc. Robotics: Science and Systems (RSS)}, 2023.

\bibitem{black2024pi0}
K.~Black, N.~Brown, D.~Driess, A.~Esmail, M.~Equi, C.~Finn, N.~Fusai, L.~Groom,
  K.~Hausman, B.~Ichter, S.~Jakubczak, T.~Jones, L.~Ke, S.~Levine, A.~Li-Bell,
  M.~Mothukuri, S.~Nair, K.~Pertsch, L.~X. Shi, J.~Tanner, Q.~Vuong,
  A.~Walling, H.~Wang, and U.~Zhilinsky, ``$\pi_0$: A vision-language-action
  flow model for general robot control,'' \emph{arXiv preprint
  arXiv:2410.24164}, 2024.

\bibitem{black2025rtc}
K.~Black, M.~Y. Galliker, and S.~Levine, ``Real-time execution of action
  chunking flow policies,'' in \emph{Advances in Neural Information Processing
  Systems (NeurIPS)}, 2025, arXiv:2506.07339.

\bibitem{zhu2026deflect}
Y.~Zhu, Y.~Chen, Z.~Yang, Y.~Hu, and X.~Chen, ``{DEFLECT}: Temporal
  counterfactual preference learning for delay-robust asynchronous {VLAs},''
  \emph{arXiv preprint arXiv:2605.19294}, 2026.

\bibitem{zhao2026dehp}
Y.~Zhao, M.~Bogdanovic, A.~Sohal, L.~Tao, K.~Darvish, A.~Aspuru-Guzik,
  F.~Shkurti, and A.~Garg, ``Dynamic execution horizon prediction for
  chunk-based robot policies,'' \emph{arXiv preprint arXiv:2606.11408}, 2026.

\bibitem{chen2026dreamchunk}
W.~Chen, K.~Zhang, C.~Lin, Z.~Zhang, Y.~She, Y.~Liu, R.~A. Yeh, S.~Mou, and
  Y.~Gu, ``{DREAM-Chunk}: Reactive action chunking with latent world model,''
  \emph{arXiv preprint arXiv:2606.18589}, 2026.

\bibitem{sun2019robust}
Z.~Sun, L.~Dai, K.~Liu, D.~V. Dimarogonas, and Y.~Xia, ``Robust self-triggered
  mpc with adaptive prediction horizon for perturbed nonlinear systems,''
  \emph{IEEE Transactions on Automatic Control}, vol.~64, no.~11, pp.
  4780--4787, 2019.

\bibitem{magdici2017horizon}
S.~Magdici, Z.~Ye, and M.~Althoff, ``Determining the maximum time horizon for
  vehicles to safely follow a trajectory,'' in \emph{Proc. IEEE Int. Conf.
  Intelligent Transportation Systems (ITSC)}, 2017, pp. 1--7.

\bibitem{dean1988timedependent}
T.~L. Dean and M.~Boddy, ``An analysis of time-dependent planning,'' in
  \emph{Proc. AAAI Conf. Artificial Intelligence}, 1988, pp. 49--54.

\bibitem{zilberstein1996anytime}
S.~Zilberstein, ``Using anytime algorithms in intelligent systems,'' \emph{AI
  Magazine}, vol.~17, no.~3, pp. 73--83, 1996.

\bibitem{callaway2018bmps}
F.~Callaway, S.~Gul, P.~M. Krueger, T.~L. Griffiths, and F.~Lieder, ``Learning
  to select computations,'' in \emph{Proc. Conf. Uncertainty in Artificial
  Intelligence (UAI)}, 2018.

\bibitem{muppidi2026budgets}
A.~Muppidi, F.~Darwish, D.~Cope, J.~F. Henriques, and J.~N. Foerster, ``Finding
  the time to think: Learning planning budgets in real-time {RL},'' \emph{arXiv
  preprint arXiv:2606.26463}, 2026.

\bibitem{sutton1999between}
R.~S. Sutton, D.~Precup, and S.~Singh, ``Between mdps and semi-mdps: A
  framework for temporal abstraction in reinforcement learning,''
  \emph{Artificial intelligence}, vol. 112, no. 1-2, pp. 181--211, 1999.

\bibitem{kuba2022mirror}
J.~G. Kuba, C.~Schroeder~de Witt, and J.~Foerster, ``Mirror learning: A
  unifying framework of policy optimisation,'' in \emph{Proc. Int. Conf.
  Machine Learning (ICML)}, ser. PMLR, vol. 162, 2022.

\bibitem{blumenkamp2024cambridge}
J.~Blumenkamp, A.~Shankar, M.~Bettini, J.~Bird, and A.~Prorok, ``The cambridge
  robomaster: An agile multi-robot research platform,'' in \emph{International
  Symposium on Distributed Autonomous Robotic Systems}.\hskip 1em plus 0.5em
  minus 0.4em\relax Springer, 2024, pp. 439--456.

\bibitem{todorov2012mujoco}
E.~Todorov, T.~Erez, and Y.~Tassa, ``Mujoco: A physics engine for model-based
  control,'' in \emph{2012 IEEE/RSJ international conference on intelligent
  robots and systems}.\hskip 1em plus 0.5em minus 0.4em\relax IEEE, 2012, pp.
  5026--5033.

\bibitem{kurtz2024hydrax}
V.~Kurtz, ``Hydrax: Sampling-based model predictive control on gpu with jax and
  mujoco mjx,'' 2024, https://github.com/vincekurtz/hydrax.

\end{thebibliography}
